\documentclass{article}[emulateapj.cls]

\usepackage{times}  
\usepackage{ulem}
\usepackage{helvet}  
\usepackage{courier}  
\usepackage[hyphens]{url}  
\usepackage{graphicx} 
\usepackage{booktabs}       
\usepackage[square,sort,comma,numbers]{natbib}

\usepackage{natbib}
\usepackage{color}
\usepackage[english]{babel}
\usepackage{amsfonts}
\usepackage{amsmath}
\usepackage{amssymb}
\usepackage{amsthm}
\usepackage{bm}
\usepackage{mathtools}
\usepackage{thmtools} 
\usepackage[linesnumbered,ruled,vlined]{algorithm2e}
\usepackage{algorithmic}
\usepackage{cancel}

\usepackage[letterpaper,top=2cm,bottom=2cm,left=3cm,right=3cm,marginparwidth=1.75cm]{geometry}

\usepackage{amsmath}
\usepackage{graphicx}

\usepackage{minitoc}
\doparttoc 
\faketableofcontents 
\usepackage[labelformat=simple]{subcaption}
\usepackage{wrapfig}

\usepackage{multicol, multirow}
\usepackage[ruled,vlined]{algorithm2e}

\newtheorem{theorem}{Theorem}
\newtheorem{definition}{Definition}
\newtheorem{proposition}{Proposition}

\newtheorem{lemma}{Lemma}

\newtheorem{proof*}{Proof}

\newcommand\blfootnote[1]{%
\begingroup
\renewcommand\thefootnote{}\footnote{#1}%
\addtocounter{footnote}{-1}%
\endgroup
}
\newcommand{\taskpolicy}{{}} 
\newcommand{\safepolicy}{{2, {\rm safe}}}
\newcommand{\intervenepolicy}{{\rm int}}
\newcommand{\cA}{{\mathcal A}}

\newcommand{\cR}{{\mathcal R}} 
\newcommand{\cS}{{\mathcal S}}

\SetKwInput{KwInput}{Input}                
\SetKwInput{KwOutput}{Output} 
\newtheorem{innercustomthm}{Theorem}
\newenvironment{customthm}[1]
  {\renewcommand\theinnercustomthm{#1}\innercustomthm}
  {\endinnercustomthm}

\title{DESTA: A Framework for Safe Reinforcement
Learning with Markov Games of Intervention}

\author{David Mguni$^{\dag,1}$, Usman Islam$^{1}$,  Yaqi Sun$^{2}$, Xiuling Zhang$^{3}$, Joel Jennings,\\ {Aivar Sootla$^{1}$, Changmin Yu$^{4}$, Ziyan Wang$^{1}$, Jun Wang$^{4}$, Yaodong Yang$^{5}$ }\\\\
$^1$Huawei R\&D UK,  $^2$Tsinghua University,  $^3$Zhejiang Normal University, \\  $^4$University College London, $^5$Institute for AI, Peking University. %
}
  
\begin{document}

\maketitle
\begin{abstract}
Reinforcement learning (RL) involves performing exploratory actions \blfootnote{$^\dag$Corresponding author <david.mguni@hotmail.com>. } in an unknown system. This can place a learning agent in dangerous and  potentially catastrophic system states. 
Current approaches for tackling safe learning in RL 
simultaneously trade-off safe exploration and  task fulfillment. 
In this paper, we introduce a new generation of RL solvers that learn to minimise safety violations while maximising the task reward to the extent that can be tolerated by the safe policy. Our approach introduces a novel two-player framework for safe RL called Distributive Exploration Safety Training Algorithm  (DESTA). 
The core of DESTA is a game between two adaptive agents: {\fontfamily{cmss}\selectfont Safety Agent} that is delegated the task of minimising safety violations and  {\fontfamily{cmss}\selectfont Task Agent} whose goal is to maximise the environment reward. Specifically, {\fontfamily{cmss}\selectfont Safety Agent} can selectively take control of the system at any given point to prevent safety violations while {\fontfamily{cmss}\selectfont Task Agent} is free to execute its policy at any other states. This framework enables {\fontfamily{cmss}\selectfont Safety Agent} to learn to take actions at certain states that minimise future safety violations, both during  training and testing time, 
while {\fontfamily{cmss}\selectfont Task Agent} performs actions that maximise the task performance everywhere else. 
Theoretically, we prove that DESTA converges to  stable points enabling safety violations of pretrained policies to be minimised.  
Empirically, we show DESTA's ability to augment the safety of existing policies and secondly, construct safe RL policies when the {\fontfamily{cmss}\selectfont Task Agent} and {\fontfamily{cmss}\selectfont Safety Agent} are trained concurrently. We demonstrate DESTA's superior performance against leading RL methods in \textit{Lunar Lander} and \textit{Frozen Lake} from OpenAI gym.
\end{abstract}








\color{red}









\color{black}
\section{Introduction}

Reinforcement learning (RL) is a framework that enables autonomous agents to learn complex behaviours from interactions with the environment  \citep{sutton2018reinforcement} in domains such as robotics and video games \citep{deisenroth2011learning,peng2017multiagent}. During its training phase, an RL agent explores using a \textit{trial and error approach} to determine the best actions. 
This process can lead
to the agent selecting actions that in some states may result in critical damage. For example, an aerial robot attempting to fly at high velocities can result in the helicopter crashing and subsequent permanent system failure.  Consequently, a major challenge in RL is to produce methods that solve the task \textit{and} ensure safety both during and after training. 
%
%
%

A common approach to tackle the problem of safe exploration in RL is to use a constrained Markov decision process (c-MDP) formulation \citep{altman1998constrained}. In this framework, the agent seeks to maximise a single objective subject to various safety constraints.  
Although c-MDP can be solved if the model is known, extending this formalism to settings in which the model is unknown remains a challenge \citep{chow2019lyapunov}. 
One of the main tools for tackling the c-MDP problem setting in RL is the Lagrangian approach for solving a constrained problem, that is,  solving $\max_\theta \min_\lambda f(\theta) - \lambda g(\theta)$ by gradient descent in $\lambda$ and ascent in $\theta$.
In contrast, to ensure their safety, animals exhibit a vast number of \textit{safety reflexes}: reactive intervention systems designed to override and assume control in dangerous situations to prevent injury. In this way, the procedures to maintain safety are managed by isolated systems.  One such example is the \textit{diving reflex}, a sequence of physiological responses to the threat of oxygen deprivation (asphyxiation) that overrides the body's basic behavioural and regulatory (homeostatic) systems~\citep{butler1997physiology}.  

Inspired by naturally occurring systems, in this paper, we tackle the challenge of learning safely in RL with a new two-agent framework for safe exploration and learning, DESTA. The framework entails an interdependent interaction between an agent, {\fontfamily{cmss}\selectfont Task Agent} whose policy maximises the set of environment rewards and an  RL agent, {\fontfamily{cmss}\selectfont Safety Agent} whose goal is to ensure the safety of the system. The {\fontfamily{cmss}\selectfont Safety Agent} has the authority to override the {\fontfamily{cmss}\selectfont Task Agent} and \textit{assume} control of the system and, using a \textit{deterministic policy} to avoid unsafe states while {\fontfamily{cmss}\selectfont Task Agent} can perform its actions everywhere else to maximise rewards. 
Transforming these components into a workable framework requires a formalism known as \textit{Markov games} (MGs). 
    To bridge the gap between the interventionist approach to safety solutions and RL machinery, we develop a new type of MG namely a \textit{nonzero-sum MG of interventions}. In this MG, two agents exchange control of the system to minimise safety violations while maximising the task objective. 
Our framework confers several key advantages: 
    %
    %
    %
    %
%
%

%
%
\textbf{1. Decoupled Objectives \& Safety Planning.} The tasks of maximising the environment reward and minimising safety violations are \textit{fully decoupled}. This means {\fontfamily{cmss}\selectfont Safety Agent} pursues its safety objectives without trading-off safety for environment rewards. DESTA has a nested sequential structure in which the {\fontfamily{cmss}\selectfont Safety Agent} first observes the {\fontfamily{cmss}\selectfont Task Agent}'s proposed action. Since the {\fontfamily{cmss}\selectfont Safety Agent}  performs \textit{safe planning}, it prevents the {\fontfamily{cmss}\selectfont Task Agent} from performing unsafe actions and anticipates future actions to minimise future safety violations states after and during training (see Experiment 1).
\newline\textbf{2. Safety Enhancement Tool.} An important use case for DESTA is an enhancement tool for pretrained policies. This enables DESTA to transform unsafe pretrained policies that maybe especially equipped to solve specific tasks into safe policies. Since the {\fontfamily{cmss}\selectfont Safety Agent} makes selective interventions, DESTA can preserve the ability of the base algorithm to solve the task (see Sec. \ref{sec:enhancement}).  
\newline\textbf{3. Selective (deterministic) interventions:} {\fontfamily{cmss}\selectfont Safety Agent} acts \textit{only} at states in which its action ensures the safety criteria. At states in which the safety criterion is irrelevant only actions that are relevant to the task are played. Moreover {\fontfamily{cmss}\selectfont Safety Agent} uses a \textit{deterministic policy} which eliminates inadvertent actions that may lead to safety violations (see Lunar Lander experiment).
\newline\textbf{4. Plug \& play.} Unlike various approaches in which safety contingencies are manually embedded into the policy e.g. \cite{pmlr-v37-schulman15}, DESTA accommodates any RL policy and various notions of safety. 

 We establish the following theoretical results that ensure DESTA's convergence to stable solutions:
\newline\textbf{i)} A Bellman equation for our game which is solved using a new Q learning variant (Theorem \ref{theorem:q_learning}). \newline\textbf{ii)} The conditions for the {\fontfamily{cmss}\selectfont Safety Agent} to perform interventions are characterised by a simple `obstacle condition' (Prop. \ref{prop:switching_times}) involving the {\fontfamily{cmss}\selectfont Safety Agent}'s value function and the agents' policies. \newline\textbf{iii)} Our Q learning variant converges to a solution using (linear) function approximators (Theorem \ref{primal_convergence_theorem}).


%
%
%
%
%
%
%
%
%
%

%
\textbf{Related Work.} 
%
%
%
Recent works in safe RL include assumptions from knowing the set of safe states and access to a safe policy \citep{Koller2018BSafeMPC, Berkenkamp2017SafeRL}, knowing the environment model \citep{Fisac2019, deanlqr2019}, having access to the cost function \citep{chow2019lyapunovbased}, having a continuous safety cost function \citep{cowenrivers2020samba} and using reversibility as a criterion for safety \citep{eysenbach2018leave}. 
 Constrained Policy Optimization (CPO) \citep{achiam2017constrained} extends trust region policy optimisation (TRPO) \citep{pmlr-v37-schulman15} with the aim of ensuring that a feasible policy stays within the constraints in expectation. 
 Similarly in the context of multi-agent RL \citep{yang2020overview}, \cite{gu2021multi} develops MACPO by extending multi-agent TRPO \citep{kuba2021trust} with safe constraints.
 However, the convergence  of these methods is still challenging; the learning dynamics tend to oscillate \cite{stooke2020responsive}, and the methodology does not readily accommodate general RL algorithms \citep{chow2018lyapunov}. 
 In \cite{tessler2018reward}, a reward-shaping approach is used to guide the learned policy to satisfy the constraints, however their approach provides no guarantees during the learning process.
In \cite{dalal2018safe}, a safety layer is introduced that acts on top of an RL agent's possibly unsafe actions to prevent safety violations though their framework does not deal with negative long-term consequences of an action. 
In \cite{bharadhwaj2021conservative}, a similar framework to CPO is used with a sparse binary safety signal where the Q function is overestimated to provide tuneable safety. 
Recently, \cite{stooke2020responsive} investigated the oscillation issue from a dynamical system point of view and introduced a treatment by applying a PID controller on the dual variable.  \cite{as2022constrained} proposed a Bayesian world model approach to safety, \cite{liu2022constrained} extended ``RL as inference''  framework to safety. Finally,~\cite{sootla2022saute,castellano2021reinforcement} solved RL with constraints imposed almost surely.   

To combat the limitations of the c-MDP formulation, several methods transform the original constraint to a more conservative one to ease the problem. For example \cite{el2016convex} replace the constraint cost with a conservative step-wise surrogate constraint. A significant drawback of these approaches is their conservativeness undermines task performance (the extent of the sub-optimality has yet to be characterised). Other approaches manually embed engineered safety-responses that are executed near safety-violating regions \citep{eysenbach2018leave,turchetta2020safe}. For example, in \cite{turchetta2020safe}, a safe teacher-student framework in which the teacher's objective is the value of the student's final policy, the agent is endowed with a pre-specified library of reset controls that it activates close to danger. These approaches can require time-consuming human input contrary to the goal of autonomous learning. They also do not allow the safety response to anticipate future the behaviour of the RL agent and do not perform safety planning. 
\section{Preliminaries}
%

\textbf{Reinforcement Learning (RL).} In RL, an agent sequentially selects actions to maximise its expected returns. The underlying problem is typically formalised as an MDP $\left\langle \mathcal{S},\mathcal{A},P,R,\gamma\right\rangle$ where $\mathcal{S}\subset \mathbb{R}^p$ is the set of states, $\mathcal{A}\subset \mathbb{R}^k$ is the set of actions, $P:\mathcal{S} \times \mathcal{A} \times \mathcal{S} \rightarrow [0, 1]$ is a transition probability function describing the system's dynamics, $R: \mathcal{S} \times \mathcal{A} \rightarrow \mathbb{R}$ is the reward function measuring the agent's performance and the factor $\gamma \in [0, 1)$ specifies the degree to which the agent's rewards are discounted over time \citep{sutton2018reinforcement}. At time $t\in 0,1,\ldots, $ the system is in state $s_{t} \in \mathcal{S}$ and the agent must
choose an action $a_{t} \in \mathcal{A}$ which transitions the system to a new state 
$s_{t+1} \sim P(\cdot|s_{t}, a_{t})$ and produces a reward $R(s_t, a_t)$. A policy $\pi: \mathcal{S} \times \mathcal{A} \rightarrow [0,1]$ is a probability distribution over state-action pairs where $\pi(a|s)$ represents the probability of selecting action $a\in\mathcal{A}$ in state $s\in\mathcal{S}$. The goal of an RL agent is to
find a policy $\hat{\pi}\in\Pi$ that maximises its expected returns given by the value function: $
v^{\pi}(s)=\mathbb{E}[\sum_{t=0}^\infty \gamma^tR(s_t,a_t)|a_t\sim\pi(\cdot|s_t)]$ where $\Pi$ is the agent's policy set. 

\textbf{Safety in RL.}
A key concern for RL in control and robotics settings is the idea of safety. This is handled in two main ways \citep{JMLR:v16:garcia15a}: using prior knowledge of safe states to constrain the policy during learning or modifying the objective to incorporate appropriate penalties or safety constraints.
The constrained MDP (c-MDP) framework \citep{cmdpbook} is a central formalism for tackling safety within RL. This involves maximising reward while maintaining costs within certain bounds which restricts the set of allowable policies for the MDP.
Formally, a c-MDP consists of an MDP $\left\langle \mathcal{S},\mathcal{A},P,R,\gamma\right\rangle$ and $\mathcal{C} = \{(L_i: \mathcal{S} \times\mathcal{A} \rightarrow \mathbb{R}, d_i \in \mathbb{R}) | i = 1, 2 \ldots n \}$, which is a set of safety constraint functions $\boldsymbol{L}:=(L_1,\ldots L_n)$ that the agent must satisfy and $\{d_i\}$ which describe the extent to which the constraints are allowed to be not satisfied.
Given a set of allowed policies $\Pi_C:=\{\pi\in\Pi :v^{\pi}_{L_i} \leq d_i\; \forall i=1,\ldots, n\}$  where $v_{L_i}^{\pi}(s)=\mathbb{E}[\sum_{t=0}^\infty \gamma^tL_i(s_t,a_t)|s_0=s]$, the c-MDP objective is to find a policy $\pi^\star$ such that $\pi^\star\in\arg\max_{\pi \in \Pi_C} v^{\pi}(s)$, for all $s \in\mathcal{S}$. The accumulative safety costs can be represented using hard constraints, this captures for example  avoiding subregions $\cal U \subset \cal S$. When $L_i$ is an indicator function i.e. takes values $\{0,1\}$, it is easy to see that each $v_{L_i}$ represents the accumulated probability of safety violations since $v_{L_i}^{\pi}(s)=\mathbb{E}[\sum_{t=0}^\infty \gamma^t1(s_t)|s_0=s]=\mathbb{P}(violation)$. 
%
%
%
%
\textit{Safe exploration in RL} 
seeks to address the challenge of learning an optimal policy for
a task while minimising the occurrence of safety violations (or catastrophic failures) during training in an unknown system \citep{hans2008safe}. 
Here, the aim is to keep the frequency of failure in each training episode as small as possible. 
%
%

\textbf{Markov games.} Our framework involves a system of two agents each with their individual objectives. Settings of this kind are formalised by MGs which model self-interested agents that act over time  \citep{deng2021complexity}. In the standard MG setup, the actions of \textit{both} agents influence both each agent's rewards and the system dynamics. Therefore, each agent $i\in\{1,2\}$ has its own reward function $R_i:\mathcal{S}\times(\times_{i=1}^2\mathcal{A}_i)\to\mathbb{R}$ and action set $\mathcal{A}_i$ and its goal is to maximise its \textit{own} expected returns. The system dynamics, influenced by both agents, are described by a probability kernel  $P:\mathcal{S} \times(\times_{i=1}^2\mathcal{A}_i) \times \mathcal{S} \rightarrow [0, 1]$. 
%
%

\section{Our Framework} 
    We now describe our framework which consists of two core components: firstly a game between two agents, the {\fontfamily{cmss}\selectfont Task Agent} and a second agent, the {\fontfamily{cmss}\selectfont Safety Agent} and, an impulse control component which is used by the {\fontfamily{cmss}\selectfont Safety Agent}. As we later explain, the impulse control component allows the {\fontfamily{cmss}\selectfont Safety Agent} to be selective about the set of states that it assumes control (and in doing so influence the transition dynamics and reward). With this, actions geared towards safety concerns are performed only at relevant states. This leaves the {\fontfamily{cmss}\selectfont Task Agent} to maximise the environment reward everywhere else. 
Different to the c-MDP formulation, the goal of minimising safety violations and maximising the task reward are now delegated to two individual agents that now have distinct objectives. Unlike classical MGs, using a form of control known as \textit{impulse control} \citep{mguni2018viscosity,mguni2022timing,yu2022seren} one of the agents, {\fontfamily{cmss}\selectfont Safety Agent} does not intervene at each state but assumes control at states that it decides.

Our framework is modelled by an MG $\mathcal{G}=\langle \mathcal{N},\mathcal{S},\mathcal{A},\mathcal{A}^\safepolicy,P,R_1,R_2,\gamma\rangle$ where $\mathcal{N}=\{${\fontfamily{cmss}\selectfont Safety Agent},{\fontfamily{cmss}\selectfont Task Agent}$\}$, $\mathcal{A}^\safepolicy\subseteq \mathcal{A}$ is the action set for the {\fontfamily{cmss}\selectfont Safety Agent} and $R_i:\mathcal{S}\times\mathcal{A}\times\mathcal{A}^\safepolicy\to\mathbb{R}$ are the one-step reward functions for agent $i\in\{1,2\}$ which we will define shortly. 
The transition probability $P:\mathcal{S}\times\mathcal{A}\times\mathcal{A}^\safepolicy\times\mathcal{S}\to[0,1]$ takes the state and action of both agents as inputs. The {\fontfamily{cmss}\selectfont Task Agent} and the {\fontfamily{cmss}\selectfont Safety Agent} use the Markov policies $\pi^\taskpolicy:\mathcal{S}\times\mathcal{A}\to[0,1]$ and $\pi^\safepolicy:\mathcal{S}\times\mathcal{A}^\safepolicy\to[0,1]$ which are contained in the sets $\Pi^\taskpolicy$ and $\Pi^\safepolicy\subset\Pi$  respectively. Importantly, the set $\Pi^\safepolicy\subset\Pi$ consists of \textit{deterministic policies} whereas the set $\Pi^\taskpolicy$ consists of stochastic policies. Therefore, whenever the {\fontfamily{cmss}\selectfont Safety Agent} assumes control, random exploratory  actions are switched off allowing the {\fontfamily{cmss}\selectfont Safety Agent} to exercise precise actions to avoid unsafe states. Lastly, the {\fontfamily{cmss}\selectfont Safety Agent} also has a policy $\mathfrak{g}_2:\mathcal{S}\to \{0,1\}$ which it uses to determine whether or not it should intervene. 

In this setup, whenever the {\fontfamily{cmss}\selectfont Safety Agent} decides to act, the transition dynamics are affected by only the {\fontfamily{cmss}\selectfont Safety Agent} while the {\fontfamily{cmss}\selectfont Task Agent} is allowed to affect the dynamics at all other times. Therefore the system transitions according to the probability kernel  $\boldsymbol{P}:\mathcal{S}\times\mathcal{A}\times\mathcal{A}^\safepolicy\times\mathcal{S}\to[0,1]$ 
given by: \begin{align}
\boldsymbol{P}(s',a^\taskpolicy,a^\safepolicy,s)=
P(s',a^\taskpolicy,s)\left(1-\boldsymbol{1}_{\mathcal{A}^\safepolicy}(a^\safepolicy)\right)+P(s',a^\safepolicy,s)\boldsymbol{1}_{\mathcal{A}^\safepolicy}(a^\safepolicy),
\end{align}
where $\boldsymbol{1}_{\mathcal{Y}}(y)$ is the indicator function which is $1$ whenever $y\in\mathcal{Y}$ and $0$ otherwise. 

\textbf{The Task Agent Objective} 
is to maximise its expected cumulative reward set by the environment (note that this does not include safety which is delegated to the {\fontfamily{cmss}\selectfont Safety Agent}). To construct the objective for the {\fontfamily{cmss}\selectfont Task Agent}, we begin by defining the function 
$R_1(s_t,a_t^\taskpolicy,a_t^\safepolicy)=
R(s_t,a_t^\taskpolicy)(1-\boldsymbol{1}_{\mathcal{A}^\safepolicy}(a_t^\safepolicy))+R(s,a_t^\safepolicy)\boldsymbol{1}_{\mathcal{A}^\safepolicy}(a_t^\safepolicy)$. 
The {\fontfamily{cmss}\selectfont Task Agent} seeks to maximise the following:
\begin{align}
v^{\pi^\taskpolicy,(\pi^\safepolicy,\mathfrak{g}_2)}_1(s)=\mathbb{E}\left[\sum_{t=0}^\infty\gamma^t R_1(s_t,a_t^\taskpolicy,a_t^\safepolicy)\Big|s_0\equiv s\right],   \label{p1objective}  \end{align}
where $a_t^\taskpolicy\sim\pi^\taskpolicy(\cdot|s_t)$ is {\fontfamily{cmss}\selectfont Task Agent}'s action and $a_t^\safepolicy\sim\pi^\safepolicy(\cdot|s_t)$ is an action chosen by the {\fontfamily{cmss}\selectfont Safety Agent}. 
Therefore, the reward received by {\fontfamily{cmss}\selectfont Task Agent} is $R(s_t,a_t^\safepolicy)$ whenever the {\fontfamily{cmss}\selectfont Safety Agent} decides to take an action and $R(s_t,a_t^\taskpolicy)$ otherwise. The {\fontfamily{cmss}\selectfont Task Agent} can be either a pretrained policy (which can be trained using RL or some other method) or a policy which is trained concurrently with the {\fontfamily{cmss}\selectfont Safety Agent} within the DESTA framework.
%
%

\textbf{The Safety Agent Objective} 
is to minimise safety violations during and after training. Unlike the {\fontfamily{cmss}\selectfont Task Agent}, each time the {\fontfamily{cmss}\selectfont Safety Agent} performs an action, it incurs a cost. This ensures any interventions by the {\fontfamily{cmss}\selectfont Safety Agent} are warranted by an increase in safety and the {\fontfamily{cmss}\selectfont Safety Agent} is selective about when it acts. The {\fontfamily{cmss}\selectfont Safety Agent}'s objective which it aims to maximise is:
\begin{align}
v^{\pi^{\taskpolicy},(\pi^\safepolicy,\mathfrak{g}_2)}_2(s)=\mathbb{E}\left[\sum_{t=0}^\infty\gamma^t\left(-\boldsymbol{\bar{L}}(s_t,a_t^\taskpolicy,a_t^\safepolicy)-c\boldsymbol{1}_{\mathcal{A}^\safepolicy}(a_t^\safepolicy))\right)\right].     
\end{align}
%
where $c\in\mathbb{R}$ is a fixed positive constant and the function $\boldsymbol{\bar{L}}$, which is provided by the environment, is defined by
$\boldsymbol{\bar{L}}(s_t,a_t^\taskpolicy,a_t^\safepolicy)=
\boldsymbol{L}(s_t,a_t^\taskpolicy)(1-\boldsymbol{1}_{\mathcal{A}^\safepolicy}(a_t^\safepolicy))+\boldsymbol{L}(s,a_t^\safepolicy)\boldsymbol{1}_{\mathcal{A}^\safepolicy}(a_t^\safepolicy)$ and $\boldsymbol{L}:=(L_1,\ldots L_n)$ is a set of constraint functions that indicate how much a given constraint has been violated. Each function $L_i$ can represent a (possibly binary) signal that indicates a visitation to an unsafe state.  Therefore to maximise its objective, the {\fontfamily{cmss}\selectfont Safety Agent} must determine the sequence of points at which the benefit of performing a precise action overcomes the cost of doing so. 


We now mention some important features. The task of ensuring safety is delegated to the {\fontfamily{cmss}\selectfont Safety agent} whose sole objective is minimise safety violations. The intervention cost it pays for its actions  induces selectivity about where it intervenes that is, it does so only at states at which intervening leads to an appreciable reduction in expected total safety violations. At all other states  (where the {\fontfamily{cmss}\selectfont Safety agent} sees no potential for safety violations), the {\fontfamily{cmss}\selectfont task agent} is free to play actions that deliver task rewards. A key aspect of DESTA is the {\fontfamily{cmss}\selectfont Safety agent} gets to observe the action the {\fontfamily{cmss}\selectfont task agent} \textit{would take} before it is allowed to do so. Since the {\fontfamily{cmss}\selectfont Safety agent} acts in response to the {\fontfamily{cmss}\selectfont Task Agent}, the {\fontfamily{cmss}\selectfont Safety agent} learns to anticipate the future joint actions and do \textit{planning} for maintaining safety. 

%
%
%
%

\textbf{The Safety Agent Impulse Control Mechanism}

The problem for the {\fontfamily{cmss}\selectfont Safety Agent} is to determine at which states it should assume control and which actions to take. We now describe how at a given state, using an impulse control policy \cite{mguni2018viscosity,mguni2021ligs}, how {\fontfamily{cmss}\selectfont Safety Agent} decides whether to override the {\fontfamily{cmss}\selectfont Task Agent} and its choice of intervention. At each state, the {\fontfamily{cmss}\selectfont Safety Agent} first makes a \textit{binary decision} to decide to \textit{assume control}.
An important feature of DESTA is the \textit{sequential decision process}. The policies $\pi$ and $\pi^\safepolicy$ first propose actions $a$ and $a_t^\safepolicy$ (resp.), which are observed by the policy $\mathfrak{g}_2$. The role of $\mathfrak{g}_2$ is to switch to the action suggested by $\pi^\safepolicy$ whenever the {\fontfamily{cmss}\selectfont Task Agent}'s action may incur a safety violation (at present or in future).
Unlike  \citep{eysenbach2018leave,turchetta2020safe}, our approach enables learning a safe intervention policy during training without the need to preprogram manually engineered safety responses and minimises safety violations during training unlike \citep{tessler2018reward,dalal2018safe}. Unlike  \citep{Fisac2019, deanlqr2019,chow2019lyapunovbased,fisac2019bridging}, which  require access to information which is not available without a priori knowledge of the environment, our framework does not require a priori knowledge of the model of the environment or the unsafe states. 


Denote by  $\{\tau_k\}_{k\geq 0}$ the \textit{intervention times} when the {\fontfamily{cmss}\selectfont Safety Agent} decides to take an action e.g. if the {\fontfamily{cmss}\selectfont Safety Agent} chooses first to intervene at state $s_6$ and again at $s_8$, then $\tau_1=6$ and $\tau_2=8$. Since $\tau_k=\inf\{t>\tau_{k-1}|s_{t}\in\mathcal{T},\mathfrak{g}_2(s_t)=1\}$, 
$\{\tau_k\}$ \textit{are \textbf{rules} that depend on the state} where $\mathcal{T}$ is the trajectory induced by the joint actions and kernel $P$.
%
By learning an optimal $\mathfrak{g}_2$, the {\fontfamily{cmss}\selectfont Safety Agent} learns the optimal states to perform an intervention. As we later explain, these intervention times are determined by an easy to evaluate condition on the state and  $\pi,\pi^\safepolicy$ proposals (see Prop. \ref{prop:switching_times}).

Learning to solve our system involves finding a solution in which the {\fontfamily{cmss}\selectfont Safety Agent} learns to assume control at a subset of states and minimise safety violations given the {\fontfamily{cmss}\selectfont Task Agent}'s policy while at all other states the {\fontfamily{cmss}\selectfont Task Agent} executes actions intended to maximise its task objective. 
In Sec. \ref{sec:analysis}, we prove the existence of a stable point of our MG and the convergence of our learning method to the solution. 
%
%
As we later show, the learning process for converges under our variant Q-learning method. 
\section{Use Cases for DESTA} \label{sec:enhancement}
\textbf{DESTA as a Safety Enhancement Tool.} A core usage for DESTA is to act as a framework for enhancing the safety of existing (pretrained) (Markov) policies. Used in this way, DESTA is agnostic to the training procedure that generated the underlying policy. With this DESTA can be used to transform unsafe pretrained or expert policies that may be specially equipped to solve specific tasks into safe policies by using the {\fontfamily{cmss}\selectfont Safety Agent} to make selective interventions to maintain safety at present and future states while preserving the ability of the base algorithm to apply actions elsewhere.  This confers a benefit of DESTA which is unlike various approaches in which safety contingencies are manually embedded into the policy e.g. \cite{pmlr-v37-schulman15}. DESTA permits easily plugging in any combination of RL policies e.g. model-free or model-based methods and, various notions of safety and generic expert policies. We demonstrate this in Sec. \ref{sec:experiments} by incorporating (a) a Dijkstra-based expert policy \cite{johnson1973note} as the {\fontfamily{cmss}\selectfont Task Agent} to exploit domain knowledge in a graph environment, while using the DESTA framework to handle the safety aspect and (b) a pretrained unsafe RL policy as the {\fontfamily{cmss}\selectfont Task Agent} which we make safe while preserving its ability to solve the task.

\textbf{DESTA as a Safe RL Framework.}  DESTA can be used as a framework for performing Safe RL. In this case, DESTA is used to train both a {\fontfamily{cmss}\selectfont Task Agent} and the {\fontfamily{cmss}\selectfont Safety Agent} concurrently. As we demonstrate in Sec. \ref{sec:experiments}, this leads to a policy which minimises safety violations both during and after training (through the interventions of the {\fontfamily{cmss}\selectfont Safety Agent}) while performing the underlying RL task.

Implementing both the safety policy and the task policy can be done using any off-the-shelf policy gradient methods without modifications. To summarise, we will learn three policies: the first one learning the actions of the task agent, the second learning the action of the  {\fontfamily{cmss}\selectfont safety agent}, if the  {\fontfamily{cmss}\selectfont safety agent} intervenes, and the final learning to intervene. 
We take an off-policy approach and every policy is an instance of SAC with appropriate action spaces and rewards. Since we learn off-policy every agent can collect the same triplets $s_t$, $a$, $s_{t+1}$, where $a$ is the current action (i.e., either the task action $a^\taskpolicy_t$ or the safety action $a^\safepolicy_t$). The rewards that the policies receive differ. The task policy receives $R_1 = R(s_t, a)$, the safety policy gets $R_2 = -\boldsymbol{L}(s_t, a) - c\boldsymbol{1}_{\mathcal{A}^\safepolicy}(a_t^\safepolicy)$, and the intervention policy rewards are $R^\intervenepolicy = -\boldsymbol{L}(s_t, a) - c\boldsymbol{1}_{\mathcal{A}^\safepolicy}(a_t^\safepolicy)$,
Note we chose to use the same triplets for all policies since we use off-policy algorithms, leading to efficient usage of the acquired data.

\begin{algorithm}[ht] 
\begin{algorithmic}[1] 
\STATE  {\textbf{Inputs:} Replay buffers ${\cal D}^\safepolicy = \{\emptyset\}$, ${\cal D}^\intervenepolicy = \{\emptyset\}$, \textsc{agent} - base agent for policy learning}
		\FOR{$N_{episodes}$}
		    \STATE State $s_0$
		    \FOR{$t=0,1,\ldots$}
    		    \STATE Sample a task action $a_{t}^\taskpolicy\sim\pi^\taskpolicy(\cdot|s_{t})$, a safe action $a_{t}^\safepolicy \sim \pi^2(\cdot | s_t)$, and an  intervention action $a_{t}^\intervenepolicy\sim \mathfrak{g}_2(\cdot | s_t) $ ($\in \{0, 1\}$) 
    		    \IF{$a_{t}^\intervenepolicy = 0$}
    		        \STATE Apply task action $a_{t}^\taskpolicy$ so $s_{t+1}\sim P(\cdot|a_{t}^\taskpolicy,s_t)$. Set $a = a_{t}^\taskpolicy$  
    		    \ELSIF{$a_{t}^\intervenepolicy = 1$}
    		        \STATE Apply safe action $a_{t}^\safepolicy$ so $s_{t+1}\sim P(\cdot|a_{t}^\safepolicy,s_t)$. Set  $a = a_{t}^\safepolicy$     
		        \ENDIF 
		     \STATE Receive reward $R(s_{t},a)$ and cost $\boldsymbol{L}(s_t,a)$
	        \STATE Set $R_1=R(s_t, a)$, $R_2 = -\boldsymbol{L}(s_t,a) - c(a)$, $R^\intervenepolicy = -\boldsymbol{L}(s_t,a) - c(a) a_{t}^\intervenepolicy$
	        \STATE Add the sample $(s_t, a, s_{t+1}, R_2)$ to ${\cal D}^\safepolicy$, the sample  $(s_t, a_{t}^\intervenepolicy, s_{t+1}, R^\intervenepolicy)$ to ${\cal D}^\intervenepolicy$
        	\ENDFOR
    	\STATE{\textbf{// Learn the individual policies}}
        \STATE {Update the policy $\pi^\safepolicy$ using ${\cal D}^\safepolicy$ and the base agent \textsc{agent}, the policy $\mathfrak{g}_2$ using ${\cal D}^\intervenepolicy$ and the base agent \textsc{agent}.} 
        \ENDFOR
	\caption{\textbf{D}istributive \textbf{E}xploration \textbf{S}afety \textbf{T}raining \textbf{A}lgorithm (DESTA)}
	\label{algo:Opt_reward_shap_psuedo} 
\end{algorithmic}         
\end{algorithm}
\vspace{-3mm}

\section{Theoretical Analysis of DESTA}\label{sec:analysis}


A key aspect of our framework is the presence of multiple RL processes that make decisions in a sequential order. In order to determine when to intervene, using the policy $\mathfrak{g}$ the {\fontfamily{cmss}\selectfont Safety Agent} must learn the states to allow the policy $\pi^\safepolicy$ to perform a safe action which the policy $\pi^\safepolicy$ must learn to select safe actions whenever it is allowed to execute an action. At a stable point of the learning processes the {\fontfamily{cmss}\selectfont Safety Agent} minimises safety violations while {\fontfamily{cmss}\selectfont Task Agent} maximises the environment reward. Additionally,  {\fontfamily{cmss}\selectfont Safety Agent} learns the set of states in which to perform an action to maintain safety at the current or future states.  In this section, we prove that DESTA converges to an optimal solution of the system. Central to DESTA is a Q-learning type method which is adapted to handle RL settings in which the the {\fontfamily{cmss}\selectfont Safety Agent} must also learn the optimal intervention criteria for the times $\{\tau_k\}$. 
We then extend the result to allow for (linear) function approximators. We provide a result that shows the optimal intervention times are characterised by an `obstacle condition' which can be evaluated online therefore allowing the $\mathfrak{g}_2$ to be learned online.

 We begin by defining a key object in preparation for our main results.
Given a function $Q^{\pi^{\taskpolicy},(\pi^\safepolicy,\mathfrak{g}_2)}:\mathcal{S}\times\mathcal{A}\to\mathbb{R},\;\forall\pi\in\Pi,\forall\mathfrak{g}_2,\forall s_{\tau_k}\in\mathcal{S}$, the intervention operator $\mathcal{M}^{(\pi^\safepolicy,\mathfrak{g}_2)}$ is given by $
\mathcal{M}^{(\pi^\safepolicy,\mathfrak{g}_2)}Q
^{\pi^{\taskpolicy},(\pi^\safepolicy,\mathfrak{g}_2)}(s_{\tau_k},a):=R(s_{\tau_k},a_{\tau_k})-c+\gamma\int_{\cS}ds'P(s';a_{\tau_k},s)v^{\pi^{\taskpolicy},(\pi^\safepolicy,\mathfrak{g}_2)}(s')\Big|a_{\tau_k}\sim\pi,$ where $\tau_k$ is an intervention time.
%
%
The quantity $\mathcal{M}Q_2^{\pi^{\taskpolicy},(\pi^\safepolicy,\mathfrak{g}_2)}$ measures the expected future stream of safety costs after an immediate intervention minus its cost for intervening. We define the operator $T$, $\forall s\in\cS, \forall\pi^\taskpolicy\in \Pi^\taskpolicy, \forall\pi^\safepolicy\in \Pi^\safepolicy$ by:
%
\begin{align}\nonumber&T v_2^{\pi^{\taskpolicy},(\pi^\safepolicy,\mathfrak{g}_2)}(s)\\&:=\max\Big\{\mathcal{M}^{(\pi^\safepolicy,\mathfrak{g}_2)}Q_2^{\pi^{\taskpolicy},(\pi^\safepolicy,\mathfrak{g}_2)}(s,a), R(s,a^\taskpolicy)+\gamma\int_{\mathcal{S}}ds'P(s';a_t^\taskpolicy,s)v_2^{\pi^{\taskpolicy},(\pi^\safepolicy,\mathfrak{g}_2)}(s')\Big|a_t^\taskpolicy\sim\pi^\taskpolicy(\cdot|s)\Big\}\nonumber
\end{align}
%

The Bellman operator $T$ captures the nested sequential structure of DESTA - the structure consists of an inner component with two terms: the first is the expected future stream of safety costs for the {\fontfamily{cmss}\selectfont Safety Agent} given it makes an intervention at the current state using its policy $\pi^\safepolicy$. The second term is the expected future stream of safety costs for the {\fontfamily{cmss}\selectfont Safety Agent} given no intervention and that the {\fontfamily{cmss}\selectfont Task Agent} executes the action $a_t^\taskpolicy\sim\pi^\taskpolicy(\cdot|s_t)$. Lastly, the outer component is an optimisation that chooses the least expected future stream of safety costs for the {\fontfamily{cmss}\selectfont Safety Agent} of the two possibilities.

\begin{theorem}\label{theorem:existence}
Given any $v_2^{\pi^{\taskpolicy},(\pi^\safepolicy,\mathfrak{g}_2)}:\mathcal{S}\times\mathcal{A}\to\mathbb{R}$,  the optimal value function $v_2$ is given by $\underset{k\to\infty}{\lim}T^kv_2^{\pi^{\taskpolicy},(\pi^\safepolicy,\mathfrak{g}_2)}=\underset{\hat{\pi}^\safepolicy,\hat{\mathfrak{g}}_2}{\max}v_2^{\pi^\taskpolicy,(\hat{\pi}^\safepolicy,\hat{\mathfrak{g}}_2)}, \forall \pi^\taskpolicy\in\Pi$.
\end{theorem}
The result of Theorem \ref{theorem:existence} enables the solution to the {\fontfamily{cmss}\selectfont Safety Agent} to be determined using a value iteration procedure.
Moreover, Theorem \ref{theorem:existence} enables a Q-learning approach \cite{bertsekas2012approximate} for finding the solution:

\begin{theorem}\label{theorem:q_learning}
Consider the following Q learning variant:
\begin{align*}
    &Q^{\pi^{\taskpolicy},(\pi^\safepolicy,\mathfrak{g}_2)}_{2,t+1}(s_t,a_t)=Q^{\pi^{\taskpolicy},(\pi^\safepolicy,\mathfrak{g}_2)}_{2,t}(s_t,a_t^\taskpolicy)-\alpha_t(s_t,a_t^\taskpolicy)Q^{\pi^{\taskpolicy},(\pi^\safepolicy,\mathfrak{g}_2)}_{2,t}(s_t,a_t^\taskpolicy)\nonumber
\\&\qquad+\alpha_t(s_t,a_t^\taskpolicy)\left[\max\left\{\mathcal{M}^{(\pi^\safepolicy,\mathfrak{g}_2)}Q^{\pi^{\taskpolicy},(\pi^\safepolicy,\mathfrak{g}_2)}_{2,t}(s_t,a), R(s_t,a_t)+\gamma Q^{\pi^{\taskpolicy},(\pi^\safepolicy,\mathfrak{g}_2)}_{2,t}(s',a_t^\taskpolicy)\right\}\right], 
\end{align*}
where $a_t^\taskpolicy\sim\pi^\taskpolicy(\cdot|s_t)$, then $Q_{2,t}(s)$ converges to $Q_2^\star$ with probability $1$ where $s_t,s_{t+1}\in\cS$ and $a_t\in\cA$.
\end{theorem}

We now extend the result to (linear) function approximators:

\begin{theorem}\label{primal_convergence_theorem}
Given a set of linearly independent basis functions $\Phi=\{\phi_1,\ldots,\phi_p\}$ with $\phi_k\in L_2,\forall k$. Algorithm 1 converges to a limit point $r^\star\in\mathbb{R}^p$ which is the unique solution to  $\Pi \mathfrak{F} (\Phi r^\star)=\Phi r^\star$ where
    $\mathfrak{F}\Lambda:=\hat{R}+\gamma P \max\{\mathcal{M}\Lambda,\Lambda\}$ . Moreover, $r^\star$ satisfies: $
    \left\|\Phi r^\star - Q^\star\right\|\leq (1-\gamma^2)^{-1/2}\left\|\Pi Q^\star-Q^\star\right\|$.
\end{theorem}
The theorem establishes the convergence of Algorithm 1 to a stable point with the use of linear function approximators. The second statement bounds the proximity of the convergence point by the smallest approximation error that can be achieved given the choice of basis functions. The following result characterises the {\fontfamily{cmss}\selectfont Safety Agent} policy $\mathfrak{g}_2$ and when the {\fontfamily{cmss}\selectfont Safety Agent} must intervene: 

\begin{proposition}\label{prop:switching_times}
The policy $\mathfrak{g}_2$ is given by: $\mathfrak{g}_2(s_t)=H(\mathcal{M}^{(\pi^\safepolicy,\mathfrak{g}_2)}Q_2- Q_2)(s_t,a_t)$, where $Q_2\equiv Q_2^{\pi^{\taskpolicy},(\pi^\safepolicy,\mathfrak{g}_2)}$ is the solution in Theorem \ref{theorem:existence}, $\mathcal{M}$ is the intervention operator and $H$ is the Heaviside function, moreover the intervention times 
are $\tau_k=\inf\{\tau>\tau_{k-1}|\mathcal{M}^{(\pi^\safepolicy,\mathfrak{g}_2)}Q_2= Q_2\}$. \end{proposition}
%
%
%
%
%
%
%
%
%
Prop. \ref{prop:switching_times} characterises the (categorical) distribution $\mathfrak{g}_2$. The times $\{\tau_k\}$ are determined by evaluating when $\mathcal{M}Q_2=Q_2$. The result yields a key aspect of DESTA for executing  {\fontfamily{cmss}\selectfont Safety agent}'s activations. 

While implementing the intervention policy appears to be straightforward by comparing value functions, it requires the optimal value functions in question. Furthermore, learning the intervention policy and these value functions simultaneously resulted in an unstable procedure. As a solution we propose to learn $\mathfrak{g}_2$ using an off-the-shelf policy gradient algorithm (such as TRPO, PPO or SAC). This policy is categorical with values $\{0, 1\}$ and has a reward $R^\intervenepolicy$ equal to $-\boldsymbol{L}(s_t, a) + c(a)$ if the  {\fontfamily{cmss}\selectfont safety agent} intervenes with an action $a=a^\safepolicy_t$, and $-\boldsymbol{L}(s_t, a)$ if the  {\fontfamily{cmss}\selectfont safety agent} does not intervene, i.e., the task agent applies an action $a=a^\taskpolicy_t$.

\section{Experiments} \label{sec:experiments}

We performed a series of challenges to see if {DESTA} \textbf{1)} learns to perform \textit{safe planning} \textbf{2)} learns to select the appropriate state to perform safe override interventions which avoids a trade-off between safety and task performance \textbf{3)} learns to use deterministic controls to ensure precise actions. Here we 
We compared the performance of {DESTA} with leading RL methods for safe learning: {SAC}, {PPO} \citep{schulman2017Proximal}, {Lagrangian PPO}, {TRPO} \citep{pmlr-v37-schulman15}, {Lagrangian TRPO} and {CPO} \citep{achiam2017constrained}. 
We then compared {DESTA} against these baselines on performance benchmarks in challenging high dimensional problems in Frozen lake  and Lunar Lander from Open AI gym \cite{brockman2016openai}. Further experiment details are in the Appendix. 

\textbf{Case I: Augmenting the Safety of a pretrained Task Agent policy.}
\newline We tested a method {DESTA-SAC-EXPERT} which uses a pretrained policy as the {\fontfamily{cmss}\selectfont Task Agent} that makes use of domain knowledge. In this case, we use the Dijkstra algorithm to find the next node on the shortest path from the current node to the goal, leaving the safety considerations to RL. Though this convincingly outperformed the baselines, we note that {DESTA-SAC} without domain knowledge yielded almost the same level of performance. Finally, we pretrained {SAC} on just the reward (i.e. an unsafe policy) and then plugged it in as the task agent to a training round of {DESTA-SAC}, resulting in a policy that solved the environment safely and quickly. These results validate the claim that {DESTA} is a flexible framework which augments the safety of both generic and specialised learning and control methods.

\begin{figure}[t!]
\begin{tabular}{cc}
  \includegraphics[height=3.2cm]{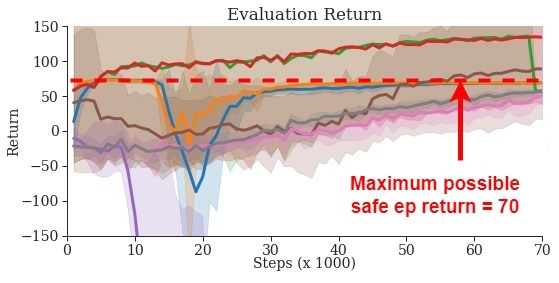} &   \includegraphics[height=3cm]{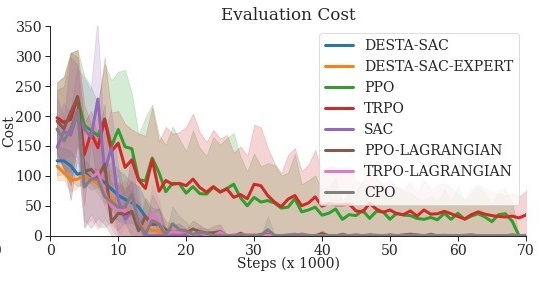} \\
(a) & (b) \\
 \includegraphics[height=3.2cm]{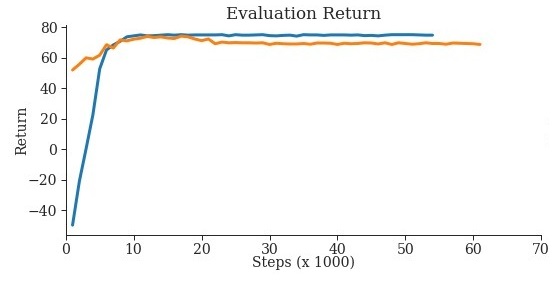} &   \includegraphics[height=3.2cm]{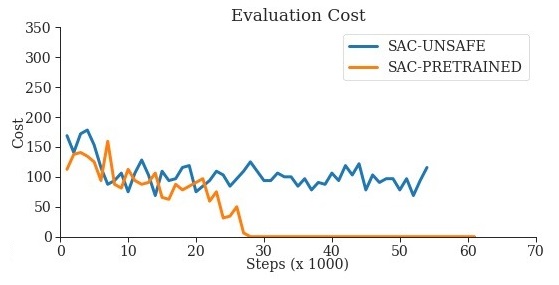} \\
(c) & (d)
\end{tabular}
\caption{\textbf{Top row:} 5 seeds evaluation of {DESTA-SAC} and baselines on Shortest Safe Route environment. (a) Average return of evaluation episodes. $70$ is the maximum safe return per episode, achieved by following the green route. Goal reward is 100 and per-step reward is -5. Cost value is 100 and cost prob in unsafe zone is 0.5. (b) Average cost of evaluation episodes. \textbf{Bottom row:} Comparison of a pretrained unsafe {SAC} policy instance ({SAC-UNSAFE}) with the resulting safe {DESTA} policy ({SAC-PRETRAINED}) (c) Evaluation return (d) Evaluation cost}
\vspace{-3
mm}
\label{fig: shortest_safe_route_results}
\end{figure}

\begin{figure}[t!]
\centering
\begin{tabular}{ccc}
  \hspace{-3 mm}\includegraphics[height=5.5cm, width = 11 cm]{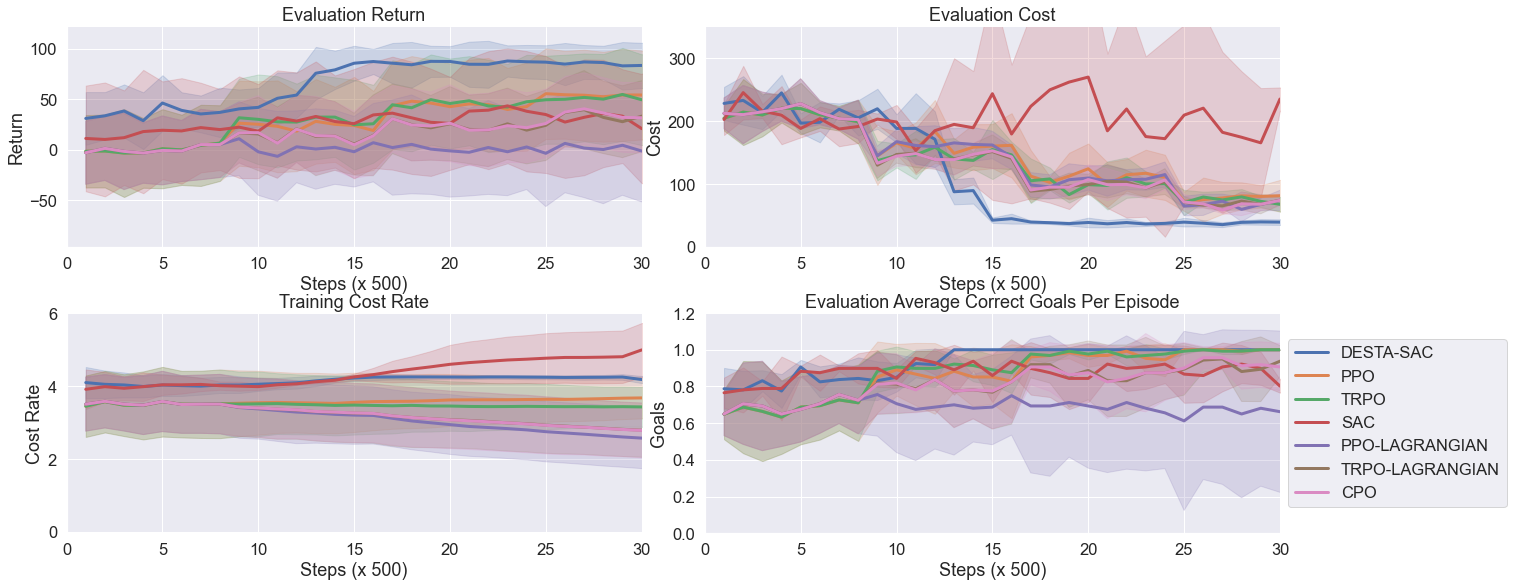} & \hspace{-3 mm}
\end{tabular}
\caption{Empirical evaluation of DESTA and baselines on Frozen Lake environment.} 
\label{fig:FrozenLake_results}
\vspace{-4
mm}
\end{figure}

\textbf{Case II: Learning the Task Agent policy and Safety Agent policy.}
\newline\textbf{Experiment 1. Safe Planning}: 
Having a dedicated agent for safety enables our method to learn to \textit{plan ahead} to minimise safety violations at future states. To demonstrate this, we designed a graph-structured environment with many routes to a shared goal (Shortest Safe Route). The agent must go from the start node to the goal node whilst avoiding the unsafe zone, where there will be a cost on each node with some (constant) probability. There is a guaranteed reward at the goal. The agent cannot go backwards so the agent's decision commits it to that path (invalid actions leave the agent unmoved). Moreover, there is a constant negative reward at each step. These features require the agent plan its route and choose the shortest route that does not incur a cost (the shortest path traverses the unsafe zone so the agent is required to make a trade-off).
We compared DESTA-SAC to {PPO}, {Lagrangian PPO}, {TRPO}, {Lagrangian TRPO}, {SAC} and {CPO}, where the (scaled) cost was deducted from the reward for {PPO}, {TRPO} and {SAC}. As shown in Fig.~\ref{fig: shortest_safe_route_results}, {DESTA-SAC} successfully learned the shortest safe route to the goal acquiring the maximum safe reward $(70)$ faster than all baselines. Other methods instead traversed unsafe zone due to the higher reward. The poor performance of {SAC} (the base learner of {DESTA-SAC}) proves DESTA is able to successfully augment safety.
%
%
%
%
%
\begin{figure}[ht]\vspace{-2.5mm}
     \centering
     \begin{subfigure}[b]{0.32\columnwidth}
     \centering
\includegraphics[height=1in]{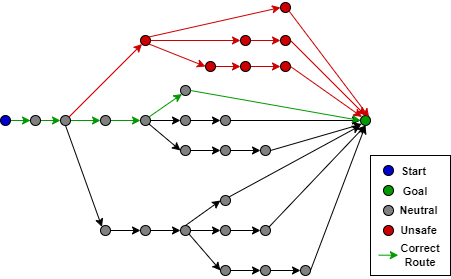}
\centering
\label{fig: shortest_safe_route_diagram}
\caption{Shortest Safe Route}
     \end{subfigure}
     \begin{subfigure}[b]{0.32\columnwidth}     
    \centering
    \includegraphics[height=1in]{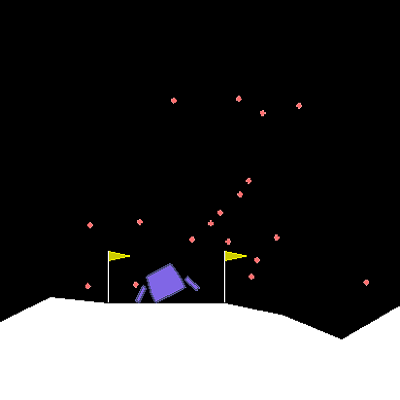}
    \caption{Lunar Lander. }\label{fig: lunar_image}
     \end{subfigure}
     \begin{subfigure}[b]{0.32\columnwidth}     
     \centering
    \includegraphics[height=1in]{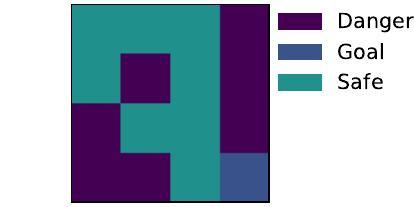}
    \caption{Frozen Lake. }\label{fig:frozen_lake}
     \end{subfigure}
    \caption{Environments and tasks: Shortest Safe Route: traverse a one-way system. Lunar lander: Land on the pad between two flags. Frozen Lake: Reach the goal while avoiding dangers.}\vspace{-3.5mm}
    \label{fig:envrionemnts}
\end{figure}
\begin{figure}[t]
\centering
\includegraphics [width=1.0\linewidth]{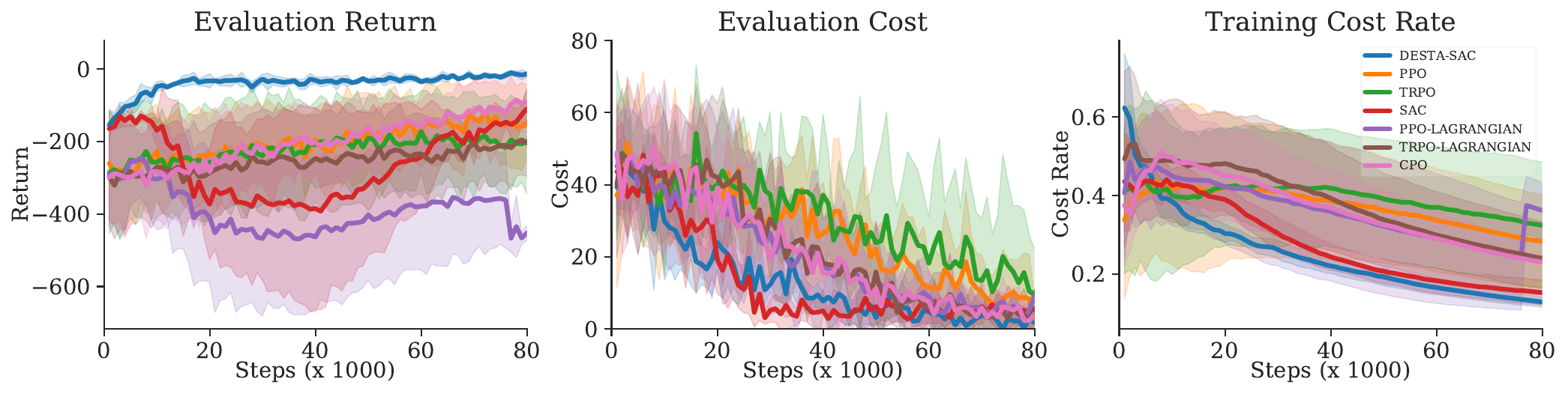}
\centering
\caption{Empirical evaluation of {DESTA} on the LunarLander task. (a) Evaluation of the agents on the LunarLander task. (b) Average cost of evaluation episodes.(c) Average training cost per step.
}
\vspace{-2.7mm}
\label{fig: lunar_eval}
\end{figure}
\newline\textbf{Experiment 2. Safe precision control using the Lunar Lander \citep{brockman2016openai}}: Since the {\fontfamily{cmss}\selectfont Safety Agent} uses determinstic controls to perform its actions, we claim DESTA is able to perform precise actions to ensure the safety of the system. 
To verify this claim, we tested DESTA's performance in the Lunar Lander environment in OpenAI gym \citep{brockman2016openai}.
As there is no strict definition of safety violation this environment~\citep{brockman2016openai}, 
we defined a safety violation to be whenever the spacecraft transitions outside a fixed horizontal threshold radius from the origin.
By introducing the safety definition, our goal was to test if DESTA can override actions that aim to exploit rewards or explore in instances when such behaviour can incur safety violations while ensuring the spacecraft is always on a near-optimal path. 
%
\newline In Fig.~\ref{fig: lunar_eval}(a) we observe {DESTA} outperforms all the baselines in terms of evaluation costs and the overall score. We observe {DESTA} enables more stable training. 
In Fig.~\ref{fig: lunar_eval}(b), we again observe that {DESTA} yields the lowest cumulative cost among all the evaluated methods while maintaining the stability of the costs across different random seeds. This indicates the {\fontfamily{cmss}\selectfont Safety Agent} has learned to better avoid the safety-violation states in comparison to the baseline methods. Fig.~\ref{fig: lunar_eval}(c) shows {DESTA} maintains the lowest cost rate throughout the training process. In this experiment, 32 evaluation episodes were run every 1000 steps and 5 seeds were used per algorithm. 
%
We note the consistent low variance across various independent training runs indicates that DESTA enables low sensitivity with respect to the randomness given different random seeds which is an issue for some RL algorithms~\citep{colas2018how}. 
\begin{wrapfigure}{r}{.38\linewidth}
\centering
\begin{tabular}{ccc}
  \hspace{-3 mm}\includegraphics[height=3cm]{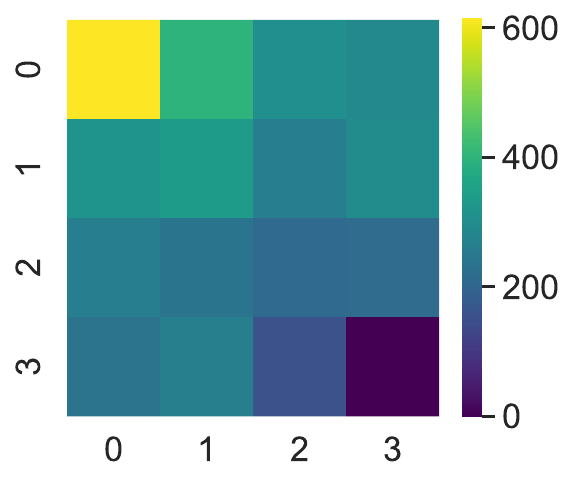} & \hspace{-3 mm}
\end{tabular}
\caption{Heatmap of intervention location during training 1000 episodes.}\label{fig:FrozenLake_heatmap}
\vspace{-3mm}
\end{wrapfigure}
\textbf{Experiment 3. Frozen lake}: In this grid world environment  (depicted in Fig.~\ref{fig:frozen_lake}) the agent's aim is to arrive at a goal as quickly as possible while avoiding unsafe and danger states. The agent receives a penalty of $-1$ for each visit to an unsafe state and a safety cost of $10$ for each visit to a dangerous state and a reward of $100$ for reaching the goal state at which point the episode terminates. The agent reward signal is observed with noise so the agent has only a $70\%$ probability of getting the reward on that grid. In this setup, there are multiple paths from the initial state to the goal --- behaving deterministically enables that agent to ensure its safety by avoiding safety violations associated with random exploration. For methods that do not decouple rewards and safety the probabilistic noise of the reward can hinder this process. Conversely, for DESTA, control of the system is transferred to safe policy which is is concerned only by safety and therefore is not affected by such noise. This allows DESTA to avoid safety violations in instances where precision is required. In this setting, at the initial state, although the both go down or right can reach the goal with the same reward, the agent will fall into a safety disaster when going down. Visualising the interventions (shown in Fig.~\ref{fig:FrozenLake_heatmap}), we observe that DESTA intervenes to encourage the agent to take immediately move right direction at the first step. Additionally, as depicted in the heatmap, DESTA becomes more active at upper left region where it intervenes to protect the agent from falling into dangerous states which may occur due to random exploration. As shown in Fig.~\ref{fig:FrozenLake_results}, DESTA solves the problem faster and than other algorithms, does so with greater stability and produces the lowest safety cost.\looseness=-1



\section{Conclusion}\label{Section:Conclusion}
We presented a novel two-player Markov game (MG) framework for solving the problem of learning safely. Our MG framework decouples the tasks of ensuring safety and maximising the task reward and assigns these tasks to a {\fontfamily{cmss}\selectfont Task Agent} and a Safety agent. This enables the {\fontfamily{cmss}\selectfont Task Agent} to learn complex behaviours to assume control at specific states to minimise safety violations and behave adaptively to the {\fontfamily{cmss}\selectfont Task Agent} whose policy is aimed at maximising the environment reward. By presenting a theoretically-grounded and high performing approach to the safe RL problem, our method opens up the applicability of RL to a range of real-world control problems with complex safety constraints.  
\clearpage
\bibliographystyle{plain}

\bibliography{main}

\begin{thebibliography}{10}

\bibitem{achiam2017constrained}
Joshua Achiam, David Held, Aviv Tamar, and Pieter Abbeel.
\newblock Constrained policy optimization, 2017.

\bibitem{altman1998constrained}
Eitan Altman.
\newblock Constrained markov decision processes with total cost criteria:
  Lagrangian approach and dual linear program.
\newblock {\em Mathematical methods of operations research}, 48(3):387--417,
  1998.

\bibitem{cmdpbook}
Eitan Altman.
\newblock {\em Constrained Markov Decision Processes}.
\newblock CRC Press, 1999.

\bibitem{as2022constrained}
Yarden As, Ilnura Usmanova, Sebastian Curi, and Andreas Krause.
\newblock Constrained policy optimization via {B}ayesian world models.
\newblock {\em arXiv preprint arXiv:2201.09802}, 2022.

\bibitem{benveniste2012adaptive}
Albert Benveniste, Michel M{\'e}tivier, and Pierre Priouret.
\newblock {\em Adaptive algorithms and stochastic approximations}, volume~22.
\newblock Springer Science \& Business Media, 2012.

\bibitem{Berkenkamp2017SafeRL}
Felix Berkenkamp, Matteo Turchetta, Angela~P. Schoellig, and Andreas Krause.
\newblock Safe model-based reinforcement learning with stability guarantees.
\newblock In {\em Neural Information Processing Systems (NeurIPS)}, 2017.

\bibitem{bertsekas2012approximate}
Dimitri~P Bertsekas.
\newblock {\em Approximate dynamic programming}.
\newblock Athena scientific Belmont, 2012.

\bibitem{bharadhwaj2021conservative}
Homanga Bharadhwaj, Aviral Kumar, Nicholas Rhinehart, Sergey Levine, Florian
  Shkurti, and Animesh Garg.
\newblock Conservative safety critics for exploration.
\newblock In {\em International Conference on Learning Representations}, 2021.

\bibitem{brockman2016openai}
Greg Brockman, Vicki Cheung, Ludwig Pettersson, Jonas Schneider, John Schulman,
  Jie Tang, and Wojciech Zaremba.
\newblock Openai gym.
\newblock {\em arXiv preprint arXiv:1606.01540}, 2016.

\bibitem{butler1997physiology}
Patrick~J Butler and David~R Jones.
\newblock Physiology of diving of birds and mammals.
\newblock {\em Physiological reviews}, 77(3):837--899, 1997.

\bibitem{castellano2021reinforcement}
Agustin Castellano, Hancheng Min, Juan Bazerque, and Enrique Mallada.
\newblock Reinforcement learning with almost sure constraints.
\newblock {\em arXiv preprint arXiv:2112.05198}, 2021.

\bibitem{chow2018lyapunov}
Yinlam Chow, Ofir Nachum, Edgar Duenez-Guzman, and Mohammad Ghavamzadeh.
\newblock A lyapunov-based approach to safe reinforcement learning.
\newblock {\em arXiv preprint arXiv:1805.07708}, 2018.

\bibitem{chow2019lyapunov}
Yinlam Chow, Ofir Nachum, Aleksandra Faust, Edgar Duenez-Guzman, and Mohammad
  Ghavamzadeh.
\newblock Lyapunov-based safe policy optimization for continuous control.
\newblock {\em arXiv preprint arXiv:1901.10031}, 2019.

\bibitem{chow2019lyapunovbased}
Yinlam Chow, Ofir Nachum, Aleksandra Faust, Edgar Duenez-Guzman, and Mohammad
  Ghavamzadeh.
\newblock Lyapunov-based safe policy optimization for continuous control, 2019.

\bibitem{colas2018how}
C.~Colas, O.~Sigaud, and P.~Y. Oudeyer.
\newblock How many random seeds? statistical power analysis in deep
  reinforcement learning experiments.
\newblock {\em arXiv preprint arXiv:1806.08295}, 2018.

\bibitem{cowenrivers2020samba}
Alexander~I. Cowen-Rivers, Daniel Palenicek, Vincent Moens, Mohammed Abdullah,
  Aivar Sootla, Jun Wang, and Haitham Ammar.
\newblock Samba: Safe model-based \& active reinforcement learning, 2020.

\bibitem{dalal2018safe}
Gal Dalal, Krishnamurthy Dvijotham, Matej Vecerik, Todd Hester, Cosmin
  Paduraru, and Yuval Tassa.
\newblock Safe exploration in continuous action spaces.
\newblock {\em arXiv preprint arXiv:1801.08757}, 2018.

\bibitem{deanlqr2019}
S.~{Dean}, S.~{Tu}, N.~{Matni}, and B.~{Recht}.
\newblock Safely learning to control the constrained linear quadratic
  regulator.
\newblock In {\em 2019 American Control Conference (ACC)}, pages 5582--5588,
  2019.

\bibitem{deisenroth2011learning}
Marc~Peter Deisenroth, Carl~Edward Rasmussen, and Dieter Fox.
\newblock Learning to control a low-cost manipulator using data-efficient
  reinforcement learning.
\newblock {\em Robotics: Science and Systems VII}, pages 57--64, 2011.

\bibitem{deng2021complexity}
Xiaotie Deng, Yuhao Li, David~Henry Mguni, Jun Wang, and Yaodong Yang.
\newblock On the complexity of computing markov perfect equilibrium in
  general-sum stochastic games.
\newblock {\em arXiv preprint arXiv:2109.01795}, 2021.

\bibitem{el2016convex}
Mahmoud El~Chamie, Yue Yu, and Beh{\c{c}}et A{\c{c}}{\i}kme{\c{s}}e.
\newblock Convex synthesis of randomized policies for controlled markov chains
  with density safety upper bound constraints.
\newblock In {\em 2016 American Control Conference (ACC)}, pages 6290--6295.
  IEEE, 2016.

\bibitem{eysenbach2018leave}
Benjamin Eysenbach, Shixiang Gu, Julian Ibarz, and Sergey Levine.
\newblock Leave no trace: Learning to reset for safe and autonomous
  reinforcement learning.
\newblock In {\em International Conference on Learning Representations}, 2018.

\bibitem{Fisac2019}
J.~F. {Fisac}, A.~K. {Akametalu}, M.~N. {Zeilinger}, S.~{Kaynama},
  J.~{Gillula}, and C.~J. {Tomlin}.
\newblock A general safety framework for learning-based control in uncertain
  robotic systems.
\newblock {\em IEEE Transactions on Automatic Control}, 64(7):2737--2752, 2019.

\bibitem{fisac2019bridging}
Jaime~F Fisac, Neil~F Lugovoy, Vicen{\c{c}} Rubies-Royo, Shromona Ghosh, and
  Claire~J Tomlin.
\newblock Bridging hamilton-jacobi safety analysis and reinforcement learning.
\newblock In {\em 2019 International Conference on Robotics and Automation
  (ICRA)}, pages 8550--8556. IEEE, 2019.

\bibitem{JMLR:v16:garcia15a}
Javier Garc{{\'i}}a, Fern, and o~Fern{{\'a}}ndez.
\newblock A comprehensive survey on safe reinforcement learning.
\newblock {\em Journal of Machine Learning Research}, 16(42):1437--1480, 2015.

\bibitem{gu2021multi}
Shangding Gu, Jakub~Grudzien Kuba, Munning Wen, Ruiqing Chen, Ziyan Wang, Zheng
  Tian, Jun Wang, Alois Knoll, and Yaodong Yang.
\newblock Multi-agent constrained policy optimisation.
\newblock {\em arXiv preprint arXiv:2110.02793}, 2021.

\bibitem{hans2008safe}
Alexander Hans, Daniel Schneega{\ss}, Anton~Maximilian Sch{\"a}fer, and Steffen
  Udluft.
\newblock Safe exploration for reinforcement learning.
\newblock In {\em ESANN}, pages 143--148. Citeseer, 2008.

\bibitem{jaakkola1994convergence}
Tommi Jaakkola, Michael~I Jordan, and Satinder~P Singh.
\newblock Convergence of stochastic iterative dynamic programming algorithms.
\newblock In {\em Advances in neural information processing systems}, pages
  703--710, 1994.

\bibitem{johnson1973note}
Donald~B Johnson.
\newblock A note on dijkstra's shortest path algorithm.
\newblock {\em Journal of the ACM (JACM)}, 20(3):385--388, 1973.

\bibitem{Koller2018BSafeMPC}
Torsten Koller, Felix Berkenkamp, Matteo Turchetta, and Andreas Krause.
\newblock Learning-based model predictive control for safe exploration.
\newblock In {\em Proc. of the Conference on Decision and Control (CDC)}, 2018.

\bibitem{kuba2021trust}
Jakub~Grudzien Kuba, Ruiqing Chen, Munning Wen, Ying Wen, Fanglei Sun, Jun
  Wang, and Yaodong Yang.
\newblock Trust region policy optimisation in multi-agent reinforcement
  learning.
\newblock {\em arXiv preprint arXiv:2109.11251}, 2021.

\bibitem{liu2022constrained}
Zuxin Liu, Zhepeng Cen, Vladislav Isenbaev, Wei Liu, Zhiwei~Steven Wu, Bo~Li,
  and Ding Zhao.
\newblock Constrained variational policy optimization for safe reinforcement
  learning.
\newblock {\em arXiv preprint arXiv:2201.11927}, 2022.

\bibitem{mguni2018viscosity}
David Mguni.
\newblock A viscosity approach to stochastic differential games of control and
  stopping involving impulsive control.
\newblock {\em arXiv preprint arXiv:1803.11432}, 2018.

\bibitem{mguni2019cutting}
David Mguni.
\newblock Cutting your losses: Learning fault-tolerant control and optimal
  stopping under adverse risk.
\newblock {\em arXiv preprint arXiv:1902.05045}, 2019.

\bibitem{mguni2022timing}
David Mguni, Aivar Sootla, Juliusz Ziomek, Oliver Slumbers, Zipeng Dai, Kun
  Shao, and Jun Wang.
\newblock Timing is everything: Learning to act selectively with costly actions
  and budgetary constraints.
\newblock {\em arXiv preprint arXiv:2205.15953}, 2022.

\bibitem{mguni2021ligs}
David~Henry Mguni, Taher Jafferjee, Jianhong Wang, Nicolas Perez-Nieves, Oliver
  Slumbers, Feifei Tong, Yang Li, Jiangcheng Zhu, Yaodong Yang, and Jun Wang.
\newblock Ligs: Learnable intrinsic-reward generation selection for multi-agent
  learning.
\newblock {\em arXiv preprint arXiv:2112.02618}, 2021.

\bibitem{peng2017multiagent}
Peng Peng, Ying Wen, Yaodong Yang, Quan Yuan, Zhenkun Tang, Haitao Long, and
  Jun Wang.
\newblock Multiagent bidirectionally-coordinated nets: Emergence of human-level
  coordination in learning to play starcraft combat games.
\newblock {\em arXiv preprint arXiv:1703.10069}, 2017.

\bibitem{pmlr-v37-schulman15}
John Schulman, Sergey Levine, Pieter Abbeel, Michael Jordan, and Philipp
  Moritz.
\newblock Trust region policy optimization.
\newblock In {\em Proceedings of the 32nd International Conference on Machine
  Learning}, pages 1889--1897, Lille, France, 07--09 Jul 2015.

\bibitem{schulman2017Proximal}
John Schulman, Filip Wolski, Prafulla Dhariwal, Alec Radford, and Oleg Klimov.
\newblock Proximal policy optimization algorithms.
\newblock {\em CoRR}, abs/1707.06347, 2017.

\bibitem{sootla2022saute}
Aivar Sootla, Alexander~I. Cowen-Rivers, Taher Jafferjee, Ziyan Wang, David
  Mguni, Jun Wang, and Haitham Bou-Ammar.
\newblock {SAUTE RL}: Almost surely safe reinforcement learning using state
  augmentation, 2022.

\bibitem{stooke2020responsive}
Adam Stooke, Joshua Achiam, and Pieter Abbeel.
\newblock Responsive safety in reinforcement learning by pid lagrangian
  methods.
\newblock In {\em International Conference on Machine Learning}, pages
  9133--9143. PMLR, 2020.

\bibitem{sutton2018reinforcement}
Richard~S Sutton and Andrew~G Barto.
\newblock {\em Reinforcement learning: An introduction}.
\newblock MIT press, 2018.

\bibitem{tessler2018reward}
Chen Tessler, Daniel~J Mankowitz, and Shie Mannor.
\newblock Reward constrained policy optimization.
\newblock {\em arXiv preprint arXiv:1805.11074}, 2018.

\bibitem{tsitsiklis1999optimal}
John~N Tsitsiklis and Benjamin Van~Roy.
\newblock Optimal stopping of markov processes: Hilbert space theory,
  approximation algorithms, and an application to pricing high-dimensional
  financial derivatives.
\newblock {\em IEEE Transactions on Automatic Control}, 44(10):1840--1851,
  1999.

\bibitem{turchetta2020safe}
Matteo Turchetta, Andrey Kolobov, Shital Shah, Andreas Krause, and Alekh
  Agarwal.
\newblock Safe reinforcement learning via curriculum induction.
\newblock {\em arXiv preprint arXiv:2006.12136}, 2020.

\bibitem{yang2020overview}
Yaodong Yang and Jun Wang.
\newblock An overview of multi-agent reinforcement learning from game
  theoretical perspective.
\newblock {\em arXiv preprint arXiv:2011.00583}, 2020.

\bibitem{yu2022seren}
Changmin Yu, David Mguni, Dong Li, Aivar Sootla, Jun Wang, and Neil Burgess.
\newblock Seren: Knowing when to explore and when to exploit.
\newblock {\em arXiv preprint arXiv:2205.15064}, 2022.

\end{thebibliography}

\clearpage
\newpage

\clearpage

\addcontentsline{toc}{section}{Appendix} 
\part{{\Large{Appendix}}} 
\parttoc
\newpage
\section{Hyperparameter Settings}

\begin{table}[ht]
    \caption{Hyperparameters for DESTA. }\label{si_tab:sac}
    \resizebox{.9\columnwidth}{!}{
    \begin{tabular}{@{}llcccc@{}}
    \toprule
    \multicolumn{2}{l}{} & \textbf{Shortest Safe Route} & \textbf{Lunar lander} & \textbf{Frozen Lake} 
    & \textbf{DynamicEnv} 
    \\    
    \midrule
    \multicolumn{1}{c}{\multirow{4}{*}{\rotatebox{90}{Runner}}} 
        & $\#$ gradient steps & $10$ & $10$ & $8$ & $8$ \\
        & $\#$ environment steps & $70k$ & $100k$ & $600k$ &$600k$\\
        & Agent frequency update & $0$ & $0$   & $4$ &$4$ \\
        & Agent batch size       & $1024$ & $1024$   & $64$ &$64$ \\
        \midrule 
\multicolumn{1}{c}{\multirow{6}{*}{\rotatebox{90}{Agents}}} 
        & Policy network dimensions  & $\{0\}$ & $\{256, 256\}$ & $\{256, 256\}$& $\{256, 256\}$\\
        & Policy networks activations  & ReLU & ReLU& ReLU & ReLU \\
        & Value network layer dims  & $\{256, 256\}$ & $\{256, 256\}$ & $\{256, 256\}$ &$\{256, 256\}$\\
        & Value networks activations  & ReLU & ReLU & ReLU &ReLU \\
        & Discount factor & $0.99$ & $0.99$ & $0.99$& $0.99$\\
        & Polyak update scale & $0$ & $0$ & $0.005$ &$0.005$ \\
        & Intervention cost & $5$ & $0.5$ & $0.1$ & $0.5$\\
        \midrule 
        \multicolumn{1}{c}{\multirow{4}{*}{\rotatebox{90}{Optimiser }}}    
        & Opt Algorithm                 & Adam            & Adam                & Adam   & Adam            \\
        & Policy learning rate  & $10^{-3}$  & $10^{-3}$ & $10^{-4}$ & $10^{-4}$\\
        & Value function learning rate    & $10^{-3}$  & $10^{-3}$ & $10^{-4}$ & $10^{-4}$\\
        & Temperature learning rate & $10^{-3}$  & $10^{-3}$ & $10^{-4}$& $10^{-4}$\\
    \bottomrule
    \end{tabular}
    }
\end{table}

\clearpage
\section{Notation \& Assumptions}\label{sec:notation_appendix}

We assume that $\mathcal{S}$ is defined on a probability space $(\Omega,\mathcal{F},\mathbb{P})$ and any $s\in\mathcal{S}$ is measurable with respect
to the Borel $\sigma$-algebra associated with $\mathbb{R}^p$. We denote the $\sigma$-algebra of events generated by $\{s_t\}_{t\geq 0}$
by $\mathcal{F}_t\subset \mathcal{F}$. In what follows, we denote by $\left( \mathcal{V},\|\|\right)$ any finite normed vector space and by $\mathcal{H}$ the set of all measurable functions. 

The results of the paper are built under the following assumptions which are standard within RL and stochastic approximation methods:

\textbf{Assumption 1.}
The stochastic process governing the system dynamics is ergodic, that is  the process is stationary and every invariant random variable of $\{s_t\}_{t\geq 0}$ is equal to a constant with probability $1$.

\textbf{Assumption 2.}
The function $R$ is in $L_2$.

\textbf{Assumption 3.}
For any positive scalar $c$, there exists a scalar $\mu_c$ such that for all $s\in\mathcal{S}$ and for any $t\in\mathbb{N}$ we have: $
    \mathbb{E}\left[1+\|s_t\|^c|s_0=s\right]\leq \mu_c(1+\|s\|^c)$.

\textbf{Assumption 4.}
There exists scalars $C_1$ and $c_1$ such that for any function $J$ satisfying $|v(s)|\leq C_2(1+\|s\|^{c_2})$ for some scalars $c_2$ and $C_2$ we have that: $
    \sum_{t=0}^\infty\left|\mathbb{E}\left[v(s_t)|s_0=s\right]-\mathbb{E}[v(s_0)]\right|\leq C_1C_2(1+\|s_t\|^{c_1c_2})$.

\textbf{Assumption 5.}
There exists scalars $c$ and $C$ such that for any $s\in\mathcal{S}$ we have that: $
    |\cR(s,\cdot)|\leq C(1+\|s\|^c)$.

\section{Proof of Technical Results}\label{sec:proofs_appendix}

We begin the analysis with some preliminary lemmata and definitions which are useful for proving the main results.

\begin{definition}{A.1}
An operator $T: \mathcal{V}\to \mathcal{V}$ is a \textbf{contraction} w.r.t a norm $\|\cdot\|$ if $\exists \lambda\in[0,1[$ such that for any $J_1,J_2\in  \mathcal{V}$ the following inequality holds: $
    \|TJ_1-TJ_2\|\leq \lambda\|J_1-J_2\|$.
    
\end{definition}

\begin{definition}{A.2}
An operator $T: \mathcal{V}\to  \mathcal{V}$ is said to be \textbf{non-expansive} if for any $J_1,J_2\in  \mathcal{V}$ the following bound holds:
    $\|TJ_1-TJ_2\|\leq \|V_1-V_2\|$.
\end{definition}

\begin{lemma} \label{max_lemma}
For any maps
$f: \mathcal{V}\to\mathbb{R},g: \mathcal{V}\to\mathbb{R}$, the following expression holds:
\begin{align}
\left\|\underset{a\in \mathcal{V}}{\max}\:f(a)-\underset{a\in \mathcal{V}}{\max}\: g(a)\right\| \leq \underset{a\in \mathcal{V}}{\max}\: \left\|f(a)-g(a)\right\|.    \label{lemma_1_basic_max_ineq}
\end{align}
\end{lemma}
\begin{proof}
We restate the proof given in \cite{mguni2019cutting}:
\begin{align}
f(a)&\leq \left\|f(a)-g(a)\right\|+g(a)\label{max_inequality_proof_start}
\\\implies
\underset{a\in \mathcal{V}}{\max}f(a)&\leq \underset{a\in \mathcal{V}}{\max}\{\left\|f(a)-g(a)\right\|+g(a)\}
\leq \underset{a\in \mathcal{V}}{\max}\left\|f(a)-g(a)\right\|+\underset{a\in \mathcal{V}}{\max}\;g(a). \label{max_inequality}
\end{align}
Deducting $\underset{a\in \mathcal{V}}{\max}\;g(a)$ from both sides of (\ref{max_inequality}) yields:
\begin{align}
    \underset{a\in \mathcal{V}}{\max}f(a)-\underset{a\in \mathcal{V}}{\max}g(a)\leq \underset{a\in \mathcal{V}}{\max}\left\|f(a)-g(a)\right\|.\label{max_inequality_result_last}
\end{align}
After reversing the roles of $f$ and $g$ and redoing steps (\ref{max_inequality_proof_start}) - (\ref{max_inequality}), we deduce the desired result since the RHS of (\ref{max_inequality_result_last}) is unchanged.
\end{proof}

\begin{lemma}{A.4}\label{non_expansive_P}
The probability transition kernel $P$ is non-expansive, that is:
\begin{align}
    \|PV_1-PV_2\|\leq \|V_1-V_2\|.
\end{align}
\end{lemma} 
\begin{proof}
The result is well-known e.g. \cite{tsitsiklis1999optimal}. We give a proof using the Tonelli-Fubini theorem and the iterated law of expectations, we have that:
\begin{align*}
&\|PJ\|^2=\mathbb{E}\left[(PJ)^2[s_0]\right]=\mathbb{E}\left(\left[\mathbb{E}\left[J[s_1]|s_0\right]\right)^2\right]
\leq \mathbb{E}\left[\mathbb{E}\left[J^2[s_1]|s_0\right]\right] 
= \mathbb{E}\left[J^2[s_1]\right]=\|J\|^2,
\end{align*}
where we have used Jensen's inequality to generate the inequality. This completes the proof.
\end{proof}

\section*{Proof of Theorem \ref{theorem:existence}}

\begin{lemma}\label{lemma:bellman_contraction}
The Bellman operator $T$ is a contraction, that is the following bound holds:
\begin{align}\nonumber
&\left\|Tv-Tv'\right\|\leq \gamma\left\|v-v'\right\|,
\end{align}
\end{lemma}
where $v,v'$ are elements of a finite normed vector space.
Lemma \ref{lemma:bellman_contraction} establishes the contraction property of the Bellman operator for the problem.

\begin{proof}
Recall we define the Bellman operator $T$ acting on a function $\Lambda:\mathcal{S}\times\mathbb{N}\to\mathbb{R}$ by
\begin{align} 
T\Lambda(s_{\tau_k}):=\max\left\{\mathcal{M}^{\pi}\Lambda(s_{\tau_k}),\left[\cR(s_{\tau_k},{a})+\gamma\underset{{a}\in\mathcal{A}}{\max}\;\sum_{s'\in\mathcal{S}}P(s';{a},s_{\tau_k})\Lambda(s',I(\tau_k))\right]\right\}\label{bellman_proof_start} 
\end{align}

For ease of exposition we use the following shorthands in the forthcoming proofs:
\begin{align*}
&\mathcal{P}^{{a}}_{ss'}=:\sum_{s'\in\mathcal{S}}P(s';{a},s), \quad\mathcal{P}^{{\pi}}_{ss'}=:\sum_{{a}\in\mathcal{A}}{\pi}({a}|s)\mathcal{P}^{{a}}_{ss'}, \quad \mathcal{R}^{{\pi}}(s_t):=\sum_{{a}_t\in\mathcal{A}}{\pi}({a}_t|s)R(s_t,a_t)
\end{align*}

To prove that $T$ is a contraction, we consider the three cases produced by \eqref{bellman_proof_start}, that is to say we prove the following statements:

i) $\qquad\qquad
\left| \cR(s_t,a_t)+\gamma\underset{{a}\in\mathcal{A}}{\max}\;\mathcal{P}^{{a}}_{s's_t}\psi(s',\cdot)-\left( \cR(s_t,a_t)+\gamma\underset{{a}\in\mathcal{A}}{\max}\;\mathcal{P}^{{a}}_{s's_t}\psi'(s',\cdot)\right)\right|\leq \gamma\left\|\psi-\psi'\right\|$

ii) $\qquad\qquad
\left\|\mathcal{M}^{\pi}\psi-\mathcal{M}^{\pi}\psi'\right\|\leq    \gamma\left\|\psi-\psi'\right\|,\qquad \qquad$
  (and hence $\mathcal{M}$ is a contraction).

iii) $\qquad\qquad
    \left\|\mathcal{M}^{\pi}\psi-\left[ \cR(\cdot,a)+\gamma\underset{{a}\in\mathcal{A}}{\max}\;\mathcal{P}^{{a}}\psi'\right]\right\|\leq \gamma\left\|\psi-\psi'\right\|.
$.

We begin by proving i).

Indeed, for any ${a}\in\mathcal{A}$ and $\forall s_t\in\mathcal{S}, \forall \theta_t,\theta_{t-1}\in \Theta, \forall s'\in\mathcal{S}$ we have that 
\begin{align*}
&\left| \cR(s_t,a_t)+\gamma\underset{{a}\in\mathcal{A}}{\max}\;\;\mathcal{P}^a_{s's_t}\psi(s',\cdot)-\left[ \cR(s_t,a_t)+\gamma\underset{{a}\in\mathcal{A}}{\max}\;\;\mathcal{P}^{{a}}_{s's_t}\psi'(s',\cdot)\right]\right|
\\&\leq \underset{{a}\in\mathcal{A}}{\max}\;\left|\gamma\mathcal{P}^{{a}}_{s's_t}\psi(s',\cdot)-\gamma\mathcal{P}^{{a}}_{s's_t}\psi'(s',\cdot)\right|
\\&\leq \gamma\left\|P\psi-P\psi'\right\|
\\&\leq \gamma\left\|\psi-\psi'\right\|,
\end{align*}
again using the fact that $P$ is non-expansive and Lemma \ref{max_lemma}.

We now prove ii).

For any $\tau\in\mathcal{F}$, define by $\tau'=\inf\{t>\tau|s_t\in A,\tau\in\mathcal{F}_t\}$. Now using the definition of $\mathcal{M}$ we have that for any $s_\tau\in\mathcal{S}$
\begin{align*}
&\left|(\mathcal{M}^{\pi}\psi-\mathcal{M}^{\pi}\psi')(s_{\tau})\right|
\\&\leq \underset{a_\tau\in \mathcal{A}}{\max}    \Bigg|\cR(s_\tau,a_\tau)+c(s_\tau,a_\tau)+\gamma\mathcal{P}^{{\pi}}_{s's_\tau}\mathcal{P}^{{a}}\psi(s_{\tau})-\left(\cR(s_\tau,a_\tau)+c(s_\tau,a_\tau)+\gamma\mathcal{P}^{{\pi}}_{s's_\tau}\mathcal{P}^{{a}}\psi'(s_{\tau})\right)\Bigg| 
\\&= \gamma\left|\mathcal{P}^{{\pi}}_{s's_\tau}\mathcal{P}^{{a}}\psi(s_{\tau})-\mathcal{P}^{{\pi}}_{s's_\tau}\mathcal{P}^{{a}}\psi'(s_{\tau})\right| 
\\&\leq \gamma\left\|P\psi-P\psi'\right\|
\\&\leq \gamma\left\|\psi-\psi'\right\|,
\end{align*}
using the fact that $P$ is non-expansive. The result can then be deduced easily by applying max on both sides.

We now prove iii). We split the proof of the statement into two cases:

\textbf{Case 1:} 
\begin{align}\mathcal{M}^{\pi}\psi(s_{\tau})-\left(\cR(s_\tau,a_\tau)+\gamma\underset{{a}\in\mathcal{A}}{\max}\;\mathcal{P}^{{a}}_{s's_\tau}\psi'(s')\right)<0.
\end{align}

We now observe the following:
\begin{align*}
&\mathcal{M}^{\pi}\psi(s_{\tau})-\cR(s_\tau,a_\tau)+\gamma\underset{{a}\in\mathcal{A}}{\max}\;\mathcal{P}^{{a}}_{s's_\tau}\psi'(s')
\\&\leq\max\left\{\cR(s_\tau,a_\tau)+\gamma\mathcal{P}^{{\pi}}_{s's_\tau}\mathcal{P}^{{a}}\psi(s'),\mathcal{M}^{\pi}\psi(s_{\tau})\right\}
-\cR(s_\tau,a_\tau)+\gamma\underset{{a}\in\mathcal{A}}{\max}\;\mathcal{P}^{{a}}_{s's_\tau}\psi'(s')
\\&\leq \Bigg|\max\left\{\cR(s_\tau,a_\tau)+\gamma\mathcal{P}^{{\pi}}_{s's_\tau}\mathcal{P}^{{a}}\psi(s'),\mathcal{M}^{\pi}\psi(s_{\tau})\right\}
-\max\left\{\cR(s_\tau,a_\tau)+\gamma\underset{{a}\in\mathcal{A}}{\max}\;\mathcal{P}^{{a}}_{s's_\tau}\psi'(s'),\mathcal{M}^{\pi}\psi(s_{\tau})\right\}
\\&+\max\left\{\cR(s_\tau,a_\tau)+\gamma\underset{{a}\in\mathcal{A}}{\max}\;\mathcal{P}^{{a}}_{s's_\tau}\psi'(s'),\mathcal{M}^{\pi}\psi(s_{\tau})\right\}
-\cR(s_\tau,a_\tau)+\gamma\underset{{a}\in\mathcal{A}}{\max}\;\mathcal{P}^{{a}}_{s's_\tau}\psi'(s')\Bigg|
\\&\leq \Bigg|\max\left\{\cR(s_\tau,a_\tau)+\gamma\underset{{a}\in\mathcal{A}}{\max}\;\mathcal{P}^{{a}}_{s's_\tau}\psi(s'),\mathcal{M}^{\pi}\psi(s_{\tau})\right\}
-\max\left\{\cR(s_\tau,a_\tau)+\gamma\underset{{a}\in\mathcal{A}}{\max}\;\mathcal{P}^{{a}}_{s's_\tau}\psi'(s'),\mathcal{M}^{\pi}\psi(s_{\tau})\right\}\Bigg|
\\&\qquad+\Bigg|\max\left\{\cR(s_\tau,a_\tau)+\gamma\underset{{a}\in\mathcal{A}}{\max}\;\mathcal{P}^{{a}}_{s's_\tau}\psi'(s'),\mathcal{M}^{\pi}\psi(s_{\tau})\right\}-\cR(s_\tau,a_\tau)+\gamma\underset{{a}\in\mathcal{A}}{\max}\;\mathcal{P}^{{a}}_{s's_\tau}\psi'(s')\Bigg|
\\&\leq \gamma\underset{a\in\mathcal{A}}{\max}\;\left|\mathcal{P}^{{\pi}}_{s's_\tau}\mathcal{P}^{{a}}\psi(s')-\mathcal{P}^{{\pi}}_{s's_\tau}\mathcal{P}^{{a}}\psi'(s')\right|
\\&\qquad+\left|\max\left\{0,\mathcal{M}^{\pi}\psi(s_{\tau})-\left(\cR(s_\tau,a_\tau)+\gamma\underset{{a}\in\mathcal{A}}{\max}\;\mathcal{P}^{{a}}_{s's_\tau}\psi'(s')\right)\right\}\right|
\\&\leq \gamma\left\|P\psi-P\psi'\right\|
\\&\leq \gamma\|\psi-\psi'\|,
\end{align*}
where we have used the fact that for any scalars $a,b,c$ we have that $
    \left|\max\{a,b\}-\max\{b,c\}\right|\leq \left|a-c\right|$ and the non-expansiveness of $P$.

\textbf{Case 2: }
\begin{align*}\mathcal{M}^{\pi}\psi(s_{\tau})-\left(\cR(s_\tau,a_\tau)+\gamma\underset{{a}\in\mathcal{A}}{\max}\;\mathcal{P}^{{a}}_{s's_\tau}\psi'(s')\right)\geq 0.
\end{align*}

For this case, first recall that for any $\tau\in\mathcal{F}$, $-c(s_\tau,a_\tau)>\lambda$ for some $\lambda >0$.
\begin{align*}
&\mathcal{M}^{\pi}\psi(s_{\tau})-\left(\cR(s_\tau,a_\tau)+\gamma\underset{{a}\in\mathcal{A}}{\max}\;\mathcal{P}^{{a}}_{s's_\tau}\psi'(s')\right)
\\&\leq \mathcal{M}^{\pi}\psi(s_{\tau})-\left(\cR(s_\tau,a_\tau)+\gamma\underset{{a}\in\mathcal{A}}{\max}\;\mathcal{P}^{{a}}_{s's_\tau}\psi'(s')\right)-c(s_\tau,a_\tau)
\\&\leq \cR(s_\tau,a_\tau)+c(s_\tau,a_\tau)+\gamma\mathcal{P}^{{\pi}}_{s's_\tau}\mathcal{P}^{{a}}\psi(s')
\\&\qquad\qquad\qquad\qquad\quad-\left(\cR(s_\tau,a_\tau)+c(s_\tau,a_\tau)+\gamma\underset{{a}\in\mathcal{A}}{\max}\;\mathcal{P}^{{a}}_{s's_\tau}\psi'(s')\right)
\\&\leq \gamma\underset{{a}\in\mathcal{A}}{\max}\;\left|\mathcal{P}^{{\pi}}_{s's_\tau}\mathcal{P}^{{a}}\left(\psi(s')-\psi'(s')\right)\right|
\\&\leq \gamma\left|\psi(s')-\psi'(s')\right|
\\&\leq \gamma\left\|\psi-\psi'\right\|,
\end{align*}
again using the fact that $P$ is non-expansive. Hence we have succeeded in showing that for any $\Lambda\in L_2$ we have that
\begin{align}
    \left\|\mathcal{M}^{\pi}\Lambda-\underset{{a}\in\mathcal{A}}{\max}\;\left[ \psi(\cdot,a)+\gamma\mathcal{P}^{{a}}\Lambda'\right]\right\|\leq \gamma\left\|\Lambda-\Lambda'\right\|.\label{off_M_bound_gen}
\end{align}
Gathering the results of the three cases gives the desired result. 
\end{proof}

To prove part ii), we make use of the following result:
\begin{theorem}[Theorem 1, pg 4 in \cite{jaakkola1994convergence}]
Let $\Xi_t(s)$ be a random process that takes values in $\mathbb{R}^n$ and given by the following:
\begin{align}
    \Xi_{t+1}(s)=\left(1-\alpha_t(s)\right)\Xi_{t}(s)\alpha_t(s)L_t(s),
\end{align}
then $\Xi_t(s)$ converges to $0$ with probability $1$ under the following conditions:
\begin{itemize}
\item[i)] $0\leq \alpha_t\leq 1, \sum_t\alpha_t=\infty$ and $\sum_t\alpha_t<\infty$
\item[ii)] $\|\mathbb{E}[L_t|\mathcal{F}_t]\|\leq \gamma \|\Xi_t\|$, with $\gamma <1$;
\item[iii)] ${\rm Var}\left[L_t|\mathcal{F}_t\right]\leq c(1+\|\Xi_t\|^2)$ for some $c>0$.
\end{itemize}
\end{theorem}

To prove the result, we show (i) - (iii) hold. Condition (i) holds by choice of learning rate. It therefore remains to prove (ii) - (iii). We first prove (ii). For this, we consider our variant of the Q-learning update rule:
\begin{align*}
Q_{t+1}(s_t,a_t)=Q_{t}&(s_t,a_t)
\\&+\alpha_t(s_t,a_t)\left[\max\left\{\mathcal{M}^{\pi}Q_t(s_t,a_t), \cR(s_t,a_t)+\gamma\underset{a'\in\mathcal{A}}{\max}\;Q_t(s_{t+1},a')\right\}-Q_{t}(s_t,a_t)\right].
\end{align*}
After subtracting $Q^\star(s_t,a_t)$ from both sides and some manipulation we obtain that:
\begin{align*}
&\Xi_{t+1}(s_t,a_t)
\\&=(1-\alpha_t(s_t,a_t))\Xi_{t}(s_t,a_t)
\\&\qquad\qquad\qquad\qquad\;\;+\alpha_t(s_t,a_t)\left[\max\left\{\mathcal{M}^{\pi}Q_t(s_t,a_t), \cR(s_t,a_t)+\gamma\underset{a'\in\mathcal{A}}{\max}\;Q_t(s_{t+1},a')\right\}-Q^\star(s_t,a_t)\right],  \end{align*}
where $\Xi_{t}(s_t,a_t):=Q_t(s_t,a_t)-Q^\star(s_t,a_t)$.

Let us now define by 
\begin{align*}
L_t(s_{\tau_k},a):=\max\left\{\mathcal{M}^{\pi}Q(s_{\tau_k},a), \cR(s_{\tau_k},a)+\gamma\underset{a'\in\mathcal{A}}{\max}\;Q(s',a')\right\}-Q^\star(s_t,a).
\end{align*}
Then
\begin{align}
\Xi_{t+1}(s_t,a_t)=(1-\alpha_t(s_t,a_t))\Xi_{t}(s_t,a_t)+\alpha_t(s_t,a_t))\left[L_t(s_{\tau_k},a)\right].   
\end{align}

We now observe that
\begin{align}\nonumber
\mathbb{E}\left[L_t(s_{\tau_k},a)|\mathcal{F}_t\right]&=\sum_{s'\in\mathcal{S}}P(s';a,s_{\tau_k})\max\left\{\mathcal{M}^{\pi}Q(s_{\tau_k},a), \cR(s_{\tau_k},a)+\gamma\underset{a'\in\mathcal{A}}{\max}\;Q(s',a')\right\}-Q^\star(s_{\tau_k},a)
\\&= T_\phi Q_t(s,a)-Q^\star(s,a). \label{expectation_L}
\end{align}
Now, using the fixed point property that implies $Q^\star=T_\phi Q^\star$, we find that
\begin{align}\nonumber
    \mathbb{E}\left[L_t(s_{\tau_k},a)|\mathcal{F}_t\right]&=T_\phi Q_t(s,a)-T_\phi Q^\star(s,a)
    \\&\leq\left\|T_\phi Q_t-T_\phi Q^\star\right\|\nonumber
    \\&\leq \gamma\left\| Q_t- Q^\star\right\|_\infty=\gamma\left\|\Xi_t\right\|_\infty.
\end{align}
using the contraction property of $T$ established in Lemma \ref{lemma:bellman_contraction}. This proves (ii).

We now prove iii), that is
\begin{align}
    {\rm Var}\left[L_t|\mathcal{F}_t\right]\leq c(1+\|\Xi_t\|^2).
\end{align}
Now by \eqref{expectation_L} we have that
\begin{align*}
  {\rm Var}\left[L_t|\mathcal{F}_t\right]&= {\rm Var}\left[\max\left\{\mathcal{M}^{\pi}Q_t(s_t,a_t), \cR(s_t,a_t)+\gamma\underset{a'\in\mathcal{A}}{\max}\;Q_t(s_{t+1},a')\right\}-Q^\star(s_t,a)\right]
  \\&= \mathbb{E}\Bigg[\Bigg(\max\left\{\mathcal{M}^{\pi}Q(s_{\tau_k},a), \cR(s_{\tau_k},a)+\gamma\underset{a'\in\mathcal{A}}{\max}\;Q(s',a')\right\}
  \\&\qquad\qquad\qquad\qquad\qquad\quad\quad\quad-Q^\star(s_t,a)-\left(T Q_t(s,a)-Q^\star(s,a)\right)\Bigg)^2\Bigg]
      \\&= \mathbb{E}\left[\left(\max\left\{\mathcal{M}^{\pi}Q(s_{\tau_k},a), \cR(s_{\tau_k},a)+\gamma\underset{a'\in\mathcal{A}}{\max}\;Q(s',a')\right\}-T Q_t(s,a)\right)^2\right]
    \\&= {\rm Var}\left[\max\left\{\mathcal{M}^{\pi}Q_t(s_t,a_t), \cR(s_t,a_t)+\gamma\underset{a'\in\mathcal{A}}{\max}\;Q_t(s_{t+1},a')\right\}-T Q_t(s,a))^2\right]
    \\&\leq c(1+\|\Xi_t\|^2),
\end{align*}
for some $c>0$ where the last line follows due to the boundedness of $Q$ (which follows from Assumptions 2 and 4). This concludes the proof of the Theorem.
\section*{Proof of Convergence with Linear Function Approximation}
First let us recall the statement of the theorem:
\begin{customthm}{3}
Algorithm 1 converges to a limit point $r^\star$ which is the unique solution to the equation:
\begin{align}
\Pi \mathfrak{F} (\Phi r^\star)=\Phi r^\star, \qquad \text{a.e.}
\end{align}
where we recall that for any test function $\Lambda \in \mathcal{V}$, the operator $\mathfrak{F}$ is defined by $
    \mathfrak{F}\Lambda:=\Theta+\gamma P \max\{\mathcal{M}\Lambda,\Lambda\}$.

Moreover, $r^\star$ satisfies the following:
\begin{align}
    \left\|\Phi r^\star - Q^\star\right\|\leq c\left\|\Pi Q^\star-Q^\star\right\|.
\end{align}
\end{customthm}

The theorem is proven using a set of results that we now establish. To this end, we first wish to prove the following bound:    
\begin{lemma}
For any $Q\in\mathcal{V}$ we have that
\begin{align}
    \left\|\mathfrak{F}Q-Q'\right\|\leq \gamma\left\|Q-Q'\right\|,
\end{align}
so that the operator $\mathfrak{F}$ is a contraction.
\end{lemma}
\begin{proof}
Recall, for any test function $\psi$ , a projection operator $\Pi$ acting $\Lambda$ is defined by the following 
\begin{align*}
\Pi \Lambda:=\underset{\bar{\Lambda}\in\{\Phi r|r\in\mathbb{R}^p\}}{\arg\min}\left\|\bar{\Lambda}-\Lambda\right\|. 
\end{align*}
Now, we first note that in the proof of Lemma \ref{lemma:bellman_contraction}, we deduced that for any $\Lambda\in L_2$ we have that
\begin{align*}
    \left\|\mathcal{M}\Lambda-\left[ \psi(\cdot,a)+\gamma\underset{a\in\mathcal{A}}{\max}\;\mathcal{P}^a\Lambda'\right]\right\|\leq \gamma\left\|\Lambda-\Lambda'\right\|,
\end{align*}
(c.f. Lemma \ref{lemma:bellman_contraction}). 

Setting $\Lambda=Q$ and $\psi=\Theta$, it can be straightforwardly deduced that for any $Q,\hat{Q}\in L_2$:
    $\left\|\mathcal{M}Q-\hat{Q}\right\|\leq \gamma\left\|Q-\hat{Q}\right\|$. Hence, using the contraction property of $\mathcal{M}$, we readily deduce the following bound:
\begin{align}\max\left\{\left\|\mathcal{M}Q-\hat{Q}\right\|,\left\|\mathcal{M}Q-\mathcal{M}\hat{Q}\right\|\right\}\leq \gamma\left\|Q-\hat{Q}\right\|,
\label{m_bound_q_twice}
\end{align}
    
We now observe that $\mathfrak{F}$ is a contraction. Indeed, since for any $Q,Q'\in L_2$ we have that:
%
%
%
\begin{align*}
\left\|\mathfrak{F}Q-\mathfrak{F}Q'\right\|&=\left\|\Theta+\gamma P \max\{\mathcal{M}Q,Q\}-\left(\Theta+\gamma P \max\{\mathcal{M}Q',Q'\}\right)\right\|
\\&=\gamma \left\|P \max\{\mathcal{M}Q,Q\}-P \max\{\mathcal{M}Q',Q'\}\right\|
\\&\leq\gamma \left\| \max\{\mathcal{M}Q,Q\}- \max\{\mathcal{M}Q',Q'\}\right\|
\\&\leq\gamma \left\| \max\{\mathcal{M}Q-\mathcal{M}Q',Q-\mathcal{M}Q',\mathcal{M}Q-Q',Q-Q'\}\right\|
\\&\leq\gamma \max\{\left\|\mathcal{M}Q-\mathcal{M}Q'\right\|,\left\|Q-\mathcal{M}Q'\right\|,\left\|\mathcal{M}Q-Q'\right\|,\left\|Q-Q'\right\|\}
\\&=\gamma\left\|Q-Q'\right\|,
\end{align*}
using \eqref{m_bound_q_twice} and again using the non-expansiveness of $P$.
\end{proof}
We next show that the following two bounds hold:
\begin{lemma}\label{projection_F_contraction_lemma}
For any $Q\in\mathcal{V}$ we have that
\begin{itemize}
    \item[i)] 
$\qquad\qquad
    \left\|\Pi \mathfrak{F}Q-\Pi \mathfrak{F}\bar{Q}\right\|\leq \gamma\left\|Q-\bar{Q}\right\|$,
    \item[ii)]$\qquad\qquad\left\|\Phi r^\star - Q^\star\right\|\leq \frac{1}{\sqrt{1-\gamma^2}}\left\|\Pi Q^\star - Q^\star\right\|$. 
\end{itemize}
\end{lemma}
\begin{proof}
The first result is straightforward since as $\Pi$ is a projection it is non-expansive and hence:
\begin{align*}
    \left\|\Pi \mathfrak{F}Q-\Pi \mathfrak{F}\bar{Q}\right\|\leq \left\| \mathfrak{F}Q-\mathfrak{F}\bar{Q}\right\|\leq \gamma \left\|Q-\bar{Q}\right\|,
\end{align*}
using the contraction property of $\mathfrak{F}$. This proves i). For ii), we note that by the orthogonality property of projections we have that $\left\langle\Phi r^\star - \Pi Q^\star,\Phi r^\star - \Pi Q^\star\right\rangle$, hence we observe that:
\begin{align*}
    \left\|\Phi r^\star - Q^\star\right\|^2&=\left\|\Phi r^\star - \Pi Q^\star\right\|^2+\left\|\Phi r^\star - \Pi Q^\star\right\|^2
\\&=\left\|\Pi \mathfrak{F}\Phi r^\star - \Pi Q^\star\right\|^2+\left\|\Phi r^\star - \Pi Q^\star\right\|^2
\\&\leq\left\|\mathfrak{F}\Phi r^\star -  Q^\star\right\|^2+\left\|\Phi r^\star - \Pi Q^\star\right\|^2
\\&=\left\|\mathfrak{F}\Phi r^\star -  \mathfrak{F}Q^\star\right\|^2+\left\|\Phi r^\star - \Pi Q^\star\right\|^2
\\&\leq\gamma^2\left\|\Phi r^\star -  Q^\star\right\|^2+\left\|\Phi r^\star - \Pi Q^\star\right\|^2,
\end{align*}
after which we readily deduce the desired result.
\end{proof}

\begin{lemma}
Define  the operator $H$ by the following: $
  HQ(z)=  \begin{cases}
			\mathcal{M}Q(z), & \text{if $\mathcal{M}Q(z)>\Phi r^\star,$}\\
            Q(z), & \text{otherwise},
		 \end{cases}$
\\and $\tilde{\mathfrak{F}}$ by: $
    \tilde{\mathfrak{F}}Q:=\Theta +\gamma PHQ$.

For any $Q,\bar{Q}\in L_2$ we have that
\begin{align}
    \left\|\tilde{\mathfrak{F}}Q-\tilde{\mathfrak{F}}\bar{Q}\right\|\leq \gamma \left\|Q-\bar{Q}\right\|
\end{align}
and hence $\tilde{\mathfrak{F}}$ is a contraction mapping.
\end{lemma}
\begin{proof}
Using \eqref{m_bound_q_twice}, we now observe that
\begin{align*}
    \left\|\tilde{\mathfrak{F}}Q-\tilde{\mathfrak{F}}\bar{Q}\right\|&=\left\|\Theta+\gamma PHQ -\left(\Theta+\gamma PH\bar{Q}\right)\right\|
\\&\leq \gamma\left\|HQ - H\bar{Q}\right\|
\\&\leq \gamma\left\|\max\left\{\mathcal{M}Q-\mathcal{M}\bar{Q},Q-\bar{Q},\mathcal{M}Q-\bar{Q},\mathcal{M}\bar{Q}-Q\right\}\right\|
\\&\leq \gamma\max\left\{\left\|\mathcal{M}Q-\mathcal{M}\bar{Q}\right\|,\left\|Q-\bar{Q}\right\|,\left\|\mathcal{M}Q-\bar{Q}\right\|,\left\|\mathcal{M}\bar{Q}-Q\right\|\right\}
\\&\leq \gamma\max\left\{\gamma\left\|Q-\bar{Q}\right\|,\left\|Q-\bar{Q}\right\|,\left\|\mathcal{M}Q-\bar{Q}\right\|,\left\|\mathcal{M}\bar{Q}-Q\right\|\right\}
\\&=\gamma\left\|Q-\bar{Q}\right\|,
\end{align*}
again using the non-expansive property of $P$.
\end{proof}
\begin{lemma}
Define by $\tilde{Q}:=\Theta+\gamma Pv^{\tilde{\pi}}$ where
\begin{align}
    v^{\tilde{\pi}}(z):= \cR(s_{\tau_k},a)+\gamma\underset{a\in\mathcal{A}}{\max}\;\sum_{s'\in\mathcal{S}}P(s';a,s_{\tau_k})\Phi r^\star(s'), \label{v_tilde_definition}
\end{align}
then $\tilde{Q}$ is a fixed point of $\tilde{\mathfrak{F}}\tilde{Q}$, that is $\tilde{\mathfrak{F}}\tilde{Q}=\tilde{Q}$. 
\end{lemma}
\begin{proof}
We begin by observing that
\begin{align*}
H\tilde{Q}(z)&=H\left(\Theta(z)+\gamma Pv^{\tilde{\pi}}\right)    
\\&= \begin{cases}
			\mathcal{M}Q(z), & \text{if $\mathcal{M}Q(z)>\Phi r^\star,$}\\
            Q(z), & \text{otherwise},
		 \end{cases}
\\&= \begin{cases}
			\mathcal{M}Q(z), & \text{if $\mathcal{M}Q(z)>\Phi r^\star,$}\\
            \Theta(z)+\gamma Pv^{\tilde{\pi}}, & \text{otherwise},
		 \end{cases}
\\&=v^{\tilde{\pi}}(z).
\end{align*}
Hence,
\begin{align}
    \tilde{\mathfrak{F}}\tilde{Q}=\Theta+\gamma PH\tilde{Q}=\Theta+\gamma Pv^{\tilde{\pi}}=\tilde{Q}. 
\end{align}
which proves the result.
\end{proof}
\begin{lemma}\label{value_difference_Q_difference}
The following bound holds:
\begin{align}
    \mathbb{E}\left[v^{\hat{\pi}}(s_0)\right]-\mathbb{E}\left[v^{\tilde{\pi}}(s_0)\right]\leq 2\left[(1-\gamma)\sqrt{(1-\gamma^2)}\right]^{-1}\left\|\Pi Q^\star -Q^\star\right\|.
\label{F_tilde_fixed_point}\end{align}
\end{lemma}
\begin{proof}

By definitions of $v^{\hat{\pi}}$ and $v^{\tilde{\pi}}$ (c.f \eqref{v_tilde_definition}) and using Jensen's inequality and the stationarity property we have that,
\begin{align}\nonumber
    \mathbb{E}\left[v^{\hat{\pi}}(s_0)\right]-\mathbb{E}\left[v^{\tilde{\pi}}(s_0)\right]&=\mathbb{E}\left[Pv^{\hat{\pi}}(s_0)\right]-\mathbb{E}\left[Pv^{\tilde{\pi}}(s_0)\right]
    \\&\leq \left|\mathbb{E}\left[Pv^{\hat{\pi}}(s_0)\right]-\mathbb{E}\left[Pv^{\tilde{\pi}}(s_0)\right]\right|\nonumber
    \\&\leq \left\|Pv^{\hat{\pi}}-Pv^{\tilde{\pi}}\right\|. \label{v_approx_intermediate_bound_P}
\end{align}
Now recall that $\tilde{Q}:=\Theta+\gamma Pv^{\tilde{\pi}}$ and $Q^\star:=\Theta+\gamma Pv^{\boldsymbol{\pi^\star}}$,  using these expressions in \eqref{v_approx_intermediate_bound_P} we find that 
\begin{align*}
    \mathbb{E}\left[v^{\hat{\pi}}(s_0)\right]-\mathbb{E}\left[v^{\tilde{\pi}}(s_0)\right]&\leq \frac{1}{\gamma}\left\|\tilde{Q}-Q^\star\right\|. \label{v_approx_q_approx_bound}
\end{align*}
Moreover, by the triangle inequality and using the fact that $\mathfrak{F}(\Phi r^\star)=\tilde{\mathfrak{F}}(\Phi r^\star)$ and that $\mathfrak{F}Q^\star=Q^\star$ and $\mathfrak{F}\tilde{Q}=\tilde{Q}$ (c.f. \eqref{F_tilde_fixed_point}) we have that
\begin{align*}
\left\|\tilde{Q}-Q^\star\right\|&\leq \left\|\tilde{Q}-\mathfrak{F}(\Phi r^\star)\right\|+\left\|Q^\star-\tilde{\mathfrak{F}}(\Phi r^\star)\right\|    
\\&\leq \gamma\left\|\tilde{Q}-\Phi r^\star\right\|+\gamma\left\|Q^\star-\Phi r^\star\right\| 
\\&\leq 2\gamma\left\|\tilde{Q}-\Phi r^\star\right\|+\gamma\left\|Q^\star-\tilde{Q}\right\|, 
\end{align*}
which gives the following bound:
\begin{align*}
\left\|\tilde{Q}-Q^\star\right\|&\leq 2\left(1-\gamma\right)^{-1}\left\|\tilde{Q}-\Phi r^\star\right\|, 
\end{align*}
from which, using Lemma \ref{projection_F_contraction_lemma}, we deduce that $
    \left\|\tilde{Q}-Q^\star\right\|\leq 2\left[(1-\gamma)\sqrt{(1-\gamma^2)}\right]^{-1}\left\|\tilde{Q}-\Phi r^\star\right\|$,
after which by \eqref{v_approx_q_approx_bound}, we finally obtain
\begin{align*}
        \mathbb{E}\left[v^{\hat{\pi}}(s_0)\right]-\mathbb{E}\left[v^{\tilde{\pi}}(s_0)\right]\leq  2\left[(1-\gamma)\sqrt{(1-\gamma^2)}\right]^{-1}\left\|\tilde{Q}-\Phi r^\star\right\|,
\end{align*}
as required.
\end{proof}

Let us rewrite the update in the following way:
\begin{align*}
    r_{t+1}=r_t+\gamma_t\Xi(w_t,r_t),
\end{align*}
where the function $\Xi:\mathbb{R}^{2d}\times \mathbb{R}^p\to\mathbb{R}^p$ is given by:
\begin{align*}
\Xi(w,r):=\phi(z)\left(\Theta(z)+\gamma\max\left\{(\Phi r) (z'),\mathcal{M}(\Phi r) (z')\right\}-(\Phi r)(z)\right),
\end{align*}
for any $w\equiv (z,z')\in\left(\mathbb{N}\times\mathcal{S}\right)^2$ where $z=(t,s)\in\mathbb{N}\times\mathcal{S}$ and $z'=(t,s')\in\mathbb{N}\times\mathcal{S}$  and for any $r\in\mathbb{R}^p$. Let us also define the function $\boldsymbol{\Xi}:\mathbb{R}^p\to\mathbb{R}^p$ by the following:
\begin{align*}
    \boldsymbol{\Xi}(r):=\mathbb{E}_{w_0\sim (\mathbb{P},\mathbb{P})}\left[\Xi(w_0,r)\right]; w_0:=(z_0,z_1).
\end{align*}
\begin{lemma}\label{iteratation_property_lemma}
The following statements hold for all $z\in \{0,1\}\times \mathcal{S}$:
\begin{itemize}
    \item[i)] $
(r-r^\star)\boldsymbol{\Xi}_k(r)<0,\qquad \forall r\neq r^\star,    
$
\item[ii)] $
\boldsymbol{\Xi}_k(r^\star)=0$.
\end{itemize}
\end{lemma}
\begin{proof}
To prove the statement, we first note that each component of $\boldsymbol{\Xi}_k(r)$ admits a representation as an inner product, indeed: 
\begin{align*}
\boldsymbol{\Xi}_k(r)&=\mathbb{E}\left[\phi_k(s_0)(\Theta(s_0)+\gamma\max\left\{\Phi r(z_1),\mathcal{M}\Phi(z_1)\right\}-(\Phi r)(s_0)\right] 
\\&=\mathbb{E}\left[\phi_k(s_0)(\Theta(s_0)+\gamma\mathbb{E}\left[\max\left\{\Phi r(z_1),\mathcal{M}\Phi(z_1)\right\}|z_0\right]-(\Phi r)(s_0)\right]
\\&=\mathbb{E}\left[\phi_k(s_0)(\Theta(s_0)+\gamma P\max\left\{\left(\Phi r,\mathcal{M}\Phi\right)\right\}(s_0)-(\Phi r)(s_0)\right]
\\&=\left\langle\phi_k,\mathfrak{F}\Phi r-\Phi r\right\rangle,
\end{align*}
using the iterated law of expectations and the definitions of $P$ and $\mathfrak{F}$.

We now are in position to prove i). Indeed, we now observe the following:
\begin{align*}
\left(r-r^\star\right)\boldsymbol{\Xi}_k(r)&=\sum_{l=1}\left(r(l)-r^\star(l)\right)\left\langle\phi_l,\mathfrak{F}\Phi r -\Phi r\right\rangle
\\&=\left\langle\Phi r -\Phi r^\star, \mathfrak{F}\Phi r -\Phi r\right\rangle
\\&=\left\langle\Phi r -\Phi r^\star, (\boldsymbol{1}-\Pi)\mathfrak{F}\Phi r+\Pi \mathfrak{F}\Phi r -\Phi r\right\rangle
\\&=\left\langle\Phi r -\Phi r^\star, \Pi \mathfrak{F}\Phi r -\Phi r\right\rangle,
\end{align*}
where in the last step we used the orthogonality of $(\boldsymbol{1}-\Pi)$. We now recall that $\Pi \mathfrak{F}\Phi r^\star=\Phi r^\star$ since $\Phi r^\star$ is a fixed point of $\Pi \mathfrak{F}$. Additionally, using Lemma \ref{projection_F_contraction_lemma} we observe that $\|\Pi \mathfrak{F}\Phi r -\Phi r^\star\| \leq \gamma \|\Phi r -\Phi r^\star\|$. With this we now find that
\begin{align*}
&\left\langle\Phi r -\Phi r^\star, \Pi \mathfrak{F}\Phi r -\Phi r\right\rangle    
\\&=\left\langle\Phi r -\Phi r^\star, (\Pi \mathfrak{F}\Phi r -\Phi r^\star)+ \Phi r^\star -\Phi r\right\rangle
\\&\leq\left\|\Phi r -\Phi r^\star\right\|\left\|\Pi \mathfrak{F}\Phi r -\Phi r^\star\right\|- \left\|\Phi r^\star -\Phi r\right\|^2
\\&\leq(\gamma -1)\left\|\Phi r^\star -\Phi r\right\|^2,
\end{align*}
which is negative since $\gamma<1$ which completes the proof of part i).

The proof of part ii) is straightforward since we readily observe that
\begin{align*}
    \boldsymbol{\Xi}_k(r^\star)= \left\langle\phi_l, \mathfrak{F}\Phi r^\star-\Phi r\right\rangle= \left\langle\phi_l, \Pi \mathfrak{F}\Phi r^\star-\Phi r\right\rangle=0,
\end{align*}
as required and from which we deduce the result.
\end{proof}
To prove the theorem, we make use of a special case of the following result:

\begin{theorem}[Th. 17, p. 239 in \cite{benveniste2012adaptive}] \label{theorem:stoch.approx.}
Consider a stochastic process $r_t:\mathbb{R}\times\{\infty\}\times\Omega\to\mathbb{R}^k$ which takes an initial value $r_0$ and evolves according to the following:
\begin{align}
    r_{t+1}=r_t+\alpha \Xi(s_t,r_t),
\end{align}
for some function $s:\mathbb{R}^{2d}\times\mathbb{R}^k\to\mathbb{R}^k$ and where the following statements hold:
\begin{enumerate}
    \item $\{s_t|t=0,1,\ldots\}$ is a stationary, ergodic Markov process taking values in $\mathbb{R}^{2d}$
    \item For any positive scalar $q$, there exists a scalar $\mu_q$ such that $\mathbb{E}\left[1+\|s_t\|^q|s\equiv s_0\right]\leq \mu_q\left(1+\|s\|^q\right)$
    \item The step size sequence satisfies the Robbins-Monro conditions, that is $\sum_{t=0}^\infty\alpha_t=\infty$ and $\sum_{t=0}^\infty\alpha^2_t<\infty$
    \item There exists scalars $c$ and $q$ such that $    \|\Xi(w,r)\|
        \leq c\left(1+\|w\|^q\right)(1+\|r\|)$
    \item There exists scalars $c$ and $q$ such that $
        \sum_{t=0}^\infty\left\|\mathbb{E}\left[\Xi(w_t,r)|z_0\equiv z\right]-\mathbb{E}\left[\Xi(w_0,r)\right]\right\|
        \leq c\left(1+\|w\|^q\right)(1+\|r\|)$
    \item There exists a scalar $c>0$ such that $
        \left\|\mathbb{E}[\Xi(w_0,r)]-\mathbb{E}[\Xi(w_0,\bar{r})]\right\|\leq c\|r-\bar{r}\| $
    \item There exists scalars $c>0$ and $q>0$ such that $
        \sum_{t=0}^\infty\left\|\mathbb{E}\left[\Xi(w_t,r)|w_0\equiv w\right]-\mathbb{E}\left[\Xi(w_0,\bar{r})\right]\right\|
        \leq c\|r-\bar{r}\|\left(1+\|w\|^q\right) $
    \item There exists some $r^\star\in\mathbb{R}^k$ such that $\boldsymbol{\Xi}(r)(r-r^\star)<0$ for all $r \neq r^\star$ and $\bar{s}(r^\star)=0$. 
\end{enumerate}
Then $r_t$ converges to $r^\star$ almost surely.
\end{theorem}

In order to apply the Theorem \ref{theorem:stoch.approx.}, we show that conditions 1 - 7 are satisfied.

\begin{proof}
Conditions 1-2 are true by assumption while condition 3 can be made true by choice of the learning rates. Therefore it remains to verify conditions 4-7 are met.   

To prove 4, we observe that
\begin{align*}
\left\|\Xi(w,r)\right\|
&=\left\|\phi(z)\left(\Theta(z)+\gamma\max\left\{(\Phi r) (z'),\mathcal{M}\Phi (z')\right\}-(\Phi r)(z)\right)\right\|
\\&\leq\left\|\phi(z)\right\|\left\|\Theta(z)+\gamma\left(\left\|\phi(z')\right\|\|r\|+\mathcal{M}\Phi (z')\right)\right\|+\left\|\phi(z)\right\|\|r\|
\\&\leq\left\|\phi(z)\right\|\left(\|\Theta(z)\|+\gamma\|\mathcal{M}\Phi (z')\|\right)+\left\|\phi(z)\right\|\left(\gamma\left\|\phi(z')\right\|+\left\|\phi(z)\right\|\right)\|r\|.
\end{align*}
Now using the definition of $\mathcal{M}$, we readily observe that $\|\mathcal{M}\Phi (z')\|\leq \| \Theta\|+\gamma\|\mathcal{P}^\pi_{s's_t}\Phi\|\leq \| \Theta\|+\gamma\|\Phi\|$ using the non-expansiveness of $P$.

Hence, we lastly deduce that
\begin{align*}
\left\|\Xi(w,r)\right\|
&\leq\left\|\phi(z)\right\|\left(\|\Theta(z)\|+\gamma\|\mathcal{M}\Phi (z')\|\right)+\left\|\phi(z)\right\|\left(\gamma\left\|\phi(z')\right\|+\left\|\phi(z)\right\|\right)\|r\|
\\&\leq\left\|\phi(z)\right\|\left(\|\Theta(z)\|+\gamma\| \Theta\|+\gamma\|\psi\|\right)+\left\|\phi(z)\right\|\left(\gamma\left\|\phi(z')\right\|+\left\|\phi(z)\right\|\right)\|r\|,
\end{align*}
we then easily deduce the result using the boundedness of $\phi,\Theta$ and $\psi$.

Now we observe the following Lipschitz condition on $\Xi$:
\begin{align*}
&\left\|\Xi(w,r)-\Xi(w,\bar{r})\right\|
\\&=\left\|\phi(z)\left(\gamma\max\left\{(\Phi r)(z'),\mathcal{M}\Phi(z')\right\}-\gamma\max\left\{(\Phi \bar{r})(z'),\mathcal{M}\Phi(z')\right\}\right)-\left((\Phi r)(z)-\Phi\bar{r}(z)\right)\right\|
\\&\leq\gamma\left\|\phi(z)\right\|\left\|\max\left\{\phi'(z') r,\mathcal{M}\Phi'(z')\right\}-\max\left\{(\phi'(z') \bar{r}),\mathcal{M}\Phi'(z')\right\}\right\|+\left\|\phi(z)\right\|\left\|\phi'(z) r-\phi(z)\bar{r}\right\|
\\&\leq\gamma\left\|\phi(z)\right\|\left\|\phi'(z') r-\phi'(z') \bar{r}\right\|+\left\|\phi(z)\right\|\left\|\phi'(z) r-\phi'(z)\bar{r}\right\|
\\&\leq \left\|\phi(z)\right\|\left(\left\|\phi(z)\right\|+ \gamma\left\|\phi(z)\right\|\left\|\phi'(z') -\phi'(z') \right\|\right)\left\|r-\bar{r}\right\|
\\&\leq c\left\|r-\bar{r}\right\|,
\end{align*}
using Cauchy-Schwarz inequality and  that for any scalars $a,b,c$ we have that $
    \left|\max\{a,b\}-\max\{b,c\}\right|\leq \left|a-c\right|$.
    
Using Assumptions 3 and 4, we therefore deduce that
\begin{align}
\sum_{t=0}^\infty\left\|\mathbb{E}\left[\Xi(w,r)-\Xi(w,\bar{r})|w_0=w\right]-\mathbb{E}\left[\Xi(w_0,r)-\Xi(w_0,\bar{r})\right\|\right]\leq c\left\|r-\bar{r}\right\|(1+\left\|w\right\|^l).
\end{align}

Part 2 is assured by Lemma \ref{projection_F_contraction_lemma} while Part 4 is assured by Lemma \ref{value_difference_Q_difference} and lastly Part 8 is assured by Lemma \ref{iteratation_property_lemma}.
This result completes the proof of Theorem \ref{theorem:existence}. 
\end{proof}

\section*{Proof of Proposition \ref{prop:switching_times}}
\begin{proof}
First let us recall that the \textit{intervention time} $\tau_k$ is defined recursively $\tau_k=\inf\{t>\tau_{k-1}|s_t\in A,\tau_k\in\mathcal{F}_t\}$ where $A=\{s\in \mathcal{S},g(s_t)=1\}$.
The proof is given by establishing a contradiction.  Therefore suppose that $\mathcal{M}^{\pi}\psi(s_{\tau_k})\leq \psi(s_{\tau_k})$ and suppose that the intervention time $\tau'_1>\tau_1$ is an optimal intervention time. Construct the $\pi'\in\Pi$ and $\tilde{\pi}\in\Pi$ policy switching times by $(\tau'_0,\tau'_1,\ldots,)$ and $(\tau'_0,\tau_1,\ldots)$ respectively.  Define by $l=\inf\{t>0;\mathcal{M}^{\pi}\psi(s_t)= \psi(s_t)\}$ and $m=\sup\{t;t<\tau'_1\}$.
By construction we have that
\begin{align*}
& \quad v^{\pi'}(s)
\\&=\mathbb{E}\left[\cR(s_{0},a_{0})+\mathbb{E}\left[\ldots+\gamma^{l-1}\mathbb{E}\left[\cR(s_{\tau_1-1},a_{\tau_1-1})+\ldots+\gamma^{m-l-1}\mathbb{E}\left[ \cR(s_{\tau'_1-1},a_{\tau'_1-1})+\gamma\mathcal{M}^{\pi^1,\pi'}v^{\pi'}(s',I(\tau'_{1}))\right]\right]\right]\right]
\\&<\mathbb{E}\left[\cR(s_{0},a_{0})+\mathbb{E}\left[\ldots+\gamma^{l-1}\mathbb{E}\left[ \cR(s_{\tau_1-1},a_{\tau_1-1})+\gamma\mathcal{M}^{\tilde{\pi}}v^{\pi'}(s_{\tau_1})\right]\right]\right]
\end{align*}
We now use the following observation 
\begin{align}
&\mathbb{E}\left[ \cR(s_{\tau_1-1},a_{\tau_1-1})+\gamma\mathcal{M}^{\tilde{\pi}}v^{\pi'}(s_{\tau_1})\right]
\\&\leq \max\left\{\mathcal{M}^{\tilde{\pi}}v^{\pi'}(s_{\tau_1}),\underset{a_{\tau_1}\in\mathcal{A}}{\max}\;\left[ \cR(s_{\tau_{k}},a_{\tau_{k}})+\gamma\sum_{s'\in\mathcal{S}}P(s';a_{\tau_1},s_{\tau_1})v^{\pi}(s')\right]\right\}.
\end{align}

Using this we deduce that
\begin{align*}
&v^{\pi'}(s)\leq\mathbb{E}\Bigg[\cR(s_{0},a_{0})+\mathbb{E}\Bigg[\ldots
\\&+\gamma^{l-1}\mathbb{E}\left[ \cR(s_{\tau_1-1},a_{\tau_1-1})+\gamma\max\left\{\mathcal{M}^{\tilde{\pi}}v^{\pi'}(s_{\tau_1}),\underset{a_{\tau_1}\in\mathcal{A}}{\max}\;\left[ \cR(s_{\tau_{k}},a_{\tau_{k}})+\gamma\sum_{s'\in\mathcal{S}}P(s';a_{\tau_1},s_{\tau_1})v^{\pi}(s')\right]\right\}\right]\Bigg]\Bigg]
\\&=\mathbb{E}\left[\cR(s_{0},a_{0})+\mathbb{E}\left[\ldots+\gamma^{l-1}\mathbb{E}\left[ \cR(s_{\tau_1-1},a_{\tau_1-1})+\gamma\left[T v^{\tilde{\pi}}\right](s_{\tau_1})\right]\right]\right]=v^{\tilde{\pi}}(s)),
\end{align*}
where the first inequality is true by assumption on $\mathcal{M}$. This is a contradiction since $\pi'$ is an optimal policy for Player 2. Using analogous reasoning, we deduce the same result for $\tau'_k<\tau_k$ after which deduce the result. Moreover, by invoking the same reasoning, we can conclude that it must be the case that $(\tau_0,\tau_1,\ldots,\tau_{k-1},\tau_k,\tau_{k+1},\ldots,)$ are the optimal switching times. 
\end{proof}

\end{document}